\documentclass[pmlr]{jmlr}

\usepackage{longtable}

\usepackage{booktabs}
\usepackage[load-configurations=version-1]{siunitx} 

\usepackage{microtype}
\usepackage{graphicx}
\usepackage{amsmath,amssymb}
\usepackage{csquotes}
\usepackage{booktabs} 
\usepackage{hyperref}
\usepackage{algorithm}
\usepackage{multirow}

\usepackage{authblk}

\usepackage[textsize=tiny]{todonotes}

\theorembodyfont{\upshape}
\theoremheaderfont{\scshape}
\theorempostheader{:}
\theoremsep{\newline}

\title[Optimizing Medical Treatments with Auto-Encoding Heuristic Search in POMDPs]{Optimizing Sequential Medical Treatments with Auto-Encoding Heuristic Search in POMDPs}

\author[1]{Luchen Li\thanks{Correspondence to Luchen Li $\langle$l.li17@imperial.ac.uk$\rangle$}}
\author[2,3]{Matthieu Komorowski}
\author[1,2]{Aldo A. Faisal}
\affil[1]{Department of Computing, Imperial College London}
\affil[2]{Department of Bioengineering, Imperial College London}
\affil[3]{Department of Surgery and Cancer, Imperial College London}

\begin{document}

\maketitle

\vspace{-1cm}
\begin{abstract}
Health-related data is noisy and stochastic in implying the true physiological states of patients, limiting information contained in single-moment observations for sequential clinical decision making. We model patient-clinician interactions as partially observable Markov decision processes (POMDPs) and optimize sequential treatment based on belief states inferred from history sequence. To facilitate inference, we build a variational generative model and boost state representation with a recurrent neural network (RNN), incorporating an auxiliary loss from sequence auto-encoding. Meanwhile, we optimize a continuous policy of drug levels with an actor-critic method where policy gradients are obtained from a stablized off-policy estimate of advantage function, with the value of belief state backed up by parallel best-first suffix trees. We exploit our methodology in optimizing dosages of vasopressor and intravenous fluid for sepsis patients using a retrospective intensive care dataset and evaluate the learned policy with off-policy policy evaluation (OPPE). The results demonstrate that modelling as POMDPs yields better performance than MDPs, and that incorporating heuristic search improves sample efficiency.
\end{abstract}

\section{Introduction}
Many recent examples \cite{gulshan2016development, esteva2017dermatologist, litjens2017survey} have demonstrated above-human performance of machine learning in classification-based diagnostics. However, the investigation in sequential treatment has not been as mature due in large part to the lack of available patient-clinician interaction trajectories. Automatic treatment optimization based on reinforcement learning (RL) has been explored in simulated patients \cite{10Ernst2006CDB, 11Bothe2013TheUO, Lowery2013}.
Public-available electronic healthcare records allow medical sequential decisions to be learned from real-world experiences \cite{8Shortreed2011ISCD, 19Asoh2013AAI, 25Lizotte2016MOMD, 12Prasad2017MVI}, even achieving human-comparable performance \cite{Komorowski2018naturemed}.

Patient-clinician interactions in these works are all modelled as Markov decision processes (MDPs). Nevertheless, what we actually can observe are patient variables, without access to the relations between them and genuine physiological states, not to mention the noisiness and inherent limitations of the apparatus undertaking the measurements, the omission of relevant factors, the incongruity of the frequencies and of the time-lags among the considered measurements, etc. In contrast, POMDP \cite{30Sondik1978TOCP} agent circumvents assumptions on any of the aforementioned phenomena by optimizing actions based on information inferred over \textit{history} observations, actions and feedback rewards. In addition, under POMDP there's no necessity to interpolate for missing data, which could just be viewed as part of the intrinsic environmental obscurity, saving incidental inaccuracies incurred by data imputation. Moreover, as we may not want to involve too much manual effort in designing rewards signals, real-world problems are more often sparse-reward tasks, introducing learning lags that harm generalization efficiency.

Algorithms in POMDPs suffer notoriously from computational complexity because the history space grows exponentially in planning horizon. Alternatively, a more memory-efficient approach \cite{35Cassandra1994AOI} is to base action selection on a probability distribution in the latent state space given history trajectory, or the belief state $b_t = P(s|\big\{ a_0, o_1, \dots, a_{t-1}, o_t \big\}, b_0)$ , where $s, ~a, ~o,~b_0$ denote the latent environment state, action, observation and an initial belief, respectively. The belief state is a sufficient statistic for history, and can be inferred incrementally. \cite{Hauskrecht2000, Tsoukalas2015, 24Nemati2016OptimalMD} build POMDP frameworks in medical context but nevertheless assume discrete observation and/or action spaces and impute trivial structures on environment dynamics.

Neural networks are used in RL to either learn nontrivial dynamics or bypass environment modelling by approximating the action-value function $Q$ in $Q$-learning, or the state-value function $V$ and a parametric policy in actor-critic \cite{Mnih2015nature, Barth2018, Horgan2018}. In addition to these function approximators, RNNs have been used in POMDPs to encode history \cite{Hausknecht2015, Zhu2017} or belief transition \cite{Igl2018}, as in POMDPs policies need to account for historical information.
However, most deep RL approaches thrive on the idealistic context of large amounts of cheap training data, unlike many real-world settings such as medical ones.

We aim to develop data-efficient POMDP agents for medical applications, and stable in the potential confrontation of significantly off-policy learning from clinicians' decisions. Specifically, each encountered belief state is evaluated through parallel suffix trees. We propose a look-ahead policy $\hat{\pi}$ to guide tree expansions that integrates the environment information embodied by a generative model into the RL agent. The belief values in the tree are subsequently back-propagated in a bottom-up fashion by Bellman update, with the values of the leaf nodes provided by current value function approximation. The value of a belief state estimated from the above simulations is closer to the true optimal value than from the value function approximation, and thus is used as the approximation target for the latter. The target policy $\pi$ is then updated through actor-critic, with policy gradient calculated from an advantage estimate. And we draw on recent literature \cite{Munos2016retrace, Wang16sampleeff, schulman2016} in gradient computation to not only assuage the bias-variance dilemma introduced by the discrepancy, i.e., \textit{off-policyness}, between $\pi$ and clinicians' behavior policy $\mu$ that actually generated the trajectories, but also enable more efficient gradient propagation despite reward sparsity.

We implement auto-encoding sequential Monte Carlo \cite{Le2018AESMC, Igl2018} to learn a generative model, with history dependencies regulated by an unrolled RNN. The inferred posterior distribution over the latent environment state space is represented by a particle filter. To pass an argument for policy learning, we aggregate information in the particle filter into a belief state. The evidence lower bound (ELBO) to the POMDP modelling and the RL loss within a mini batch are integrated to update all parameters simultaneously.

We evaluate our approach, auto-encoding heuristic search (AEHS), on a retrospective hospital database Medical Information Mart for Intensive Care Clinical Database (MIMIC-III) for optimizing sepsis treatment, learned from different numbers of observed patient variables that indicate levels of data obscurity. We demonstrate that including heuristics reduces the number of patient variables required to outperform clinicians' behavior policy, and that the POMDP is a more appropriate framework than MDP for patient trajectories.

\section{Related Work}

Earlier POMDP algorithms are built upon known model dynamics and in discrete spaces, focusing on efficient $\alpha$-vectors representations of belief value functions \cite{35Cassandra1994AOI, 33Littman1995LPP, 38Cassandra97incrementalpruning:, 40Hauskrecht2000VAP, 29Pineau2003PBVI, 39Spaan2004APPOMDP, 42Smith2004HSV, 52Ross2007AAO, Kurniawati2008SARSOP}.

For unknown environments, Bayesian approaches \cite{Ross2007Bayespomdp, DoshiVelez2015}, active learning \cite{49DOSHIVELEZ2012RLWLR} and spectral methods \cite{Azizzadenesheli2016} are explored for POMDPs to learn a model. \cite{47Doshi-velez2009TIPO} learns a set of infinite hidden Markov models and iteratively updates the weight of each. These approaches are however constrained to explicit structures of model dynamics, and are thus not scalable to larger spaces.

To handle large or continuous spaces, \cite{Coquelin2009} represents belief state with a weighted particle filter, and \cite{Silver2010POMCP} with an unweighted one, both however built upon existing environment dynamics, and omission to state uncertainty after each transition by planning over a weighted average of the particles. Moreover, represented merely as a vector random variable, their environment state might be hard to infer from history.

Owing to its capacity for processing sequences, RNNs are utilized in POMDPs to encode histories to action advantages \cite{Bakker2001}, action probabilities \cite{Wierstra2007}, features in addition to latest observation and action \cite{Zhang2015Robo}, or an intermediate layer in deep $Q$-networks (DQNs) \cite{Hausknecht2015, Zhu2017}.

However, back-propagation through time (BPTT) over long distances is notoriously susceptible to gradient vanishing/exploding. Recent works have also investigated appropriate initial states \cite{Kapturowski2019} or shortening BPTT length \cite{trinh18}. Alternatively, RNNs can be involved merely in an auxiliary task. Unsupervised auxiliary losses have been demonstrated capable of finding meaningful signals from unknown environments conducive to learning the main task \cite{Jaderberg2016}. For example, the DVRL in \cite{Igl2018} exploits auxiliary loss from sequence modelling to facilitate posterior inference, with state representation augmented by an RNN.

We propose a novel POMDP solution. Similarly to the DVRL, we conduct belief inference under a sequential variational model, and optimize actions in the inferred belief state. In the meantime, we formalize an efficient search mechanism in computing gradients for value function to improve data efficiency, and a stablized estimation of advantage function in computing those for policy to account for both policy discrepancy and reward-sparsity.

\section{Preliminary}
This section introduces recently-proposed elements that we employ to build a generative model for a sequential system and for RL online planning. For more detailed information, we refer readers to the original papers.

\subsection{Sequential Monte Carlo \& Auto-encoding}
Sequential Monte Carlo (SMC) \cite{Doucet2009SMC} is an efficient tool for performing posterior inference in structured probabilistic models of series data.

Consider a sequence of latent variables $s_{0:t}$, a sequence of observed variables $o_{1:t}$, a generative model that consists of a transition density $p_\theta (s_t|s_{0:t-1})$, an observation density $p_\theta (o_t|s_{0:t})$, an initial density $p_\theta (s_0)$\footnote{Note we sample at $t=0$ from $p_\theta (s_0)$ in accordance with the RL context instead of $q_\phi (s_0)$.}, characterized by parameters $\theta$, and an encoder parameterized by $\phi$, $q_\phi (s_t|o_{1:t}, s_{0:t-1})$. The joint likelihood is
\begin{equation}\label{eq:SMC_joint_ll}
p_\theta (s_{0:t}, o_{1:t}) = p_\theta (s_0)\prod_{\tau=1}^t p_\theta (s_\tau|s_{0:\tau-1}) p_\theta (o_\tau|s_{0:\tau})
\end{equation}
The target posterior distribution we want to perform inference over is
\begin{equation}\label{eq:SMC_posterior}
p_{\theta, \phi}(s_{0:t}|o_{1:t}) = p_\theta (s_0) \prod_{\tau=1}^t q_\phi (s_\tau|o_{1:\tau}, s_{0:\tau-1})
\end{equation}

Maintaining a set of $K$ weighted particles $\left\langle s_{0:t}^k, w_t^k \right\rangle$ at each time step, an unbiased estimator of the marginal likelihood $p_{\theta, \phi} (o_{1:t})$ can be calculated from the weights $w^{1:K}_{1:t}$
\begin{equation}\label{eq:SMC_estimator}
\hat{p}_{\theta, \phi}^\mathrm{SMC} (o_{1:t}; x_{0:t-1}^{1:K}, s_{0:t}^{1:K}) = \prod_{\tau=1}^t \big( \frac{1}{K}\sum_{k=1}^K w_\tau^k \big)
\end{equation}
when the particle filter is updated by the following procedure, i.e. mutation-selection:

1) resample index $x^k_{t-1}$ from predecessor set $x^k_{t-1}\sim w^{x^k_{t-1}}_{t-1}/\sum_{i=1}^K w^i_{t-1}$ (selection)
\vspace{-0.3cm}
\begin{equation*}
\left.\begin{aligned}
  2)& \mathrm{~sample ~new ~state~} s^k_t\sim q_\phi(\cdot|o_{1:t}, s_{0:t-1}^{x^k_{t-1}}) \\
  3)& \mathrm{~update ~particle~} s^k_{0:t}\gets (s_{0:t-1}^{x^k_{t-1}}, s^k_t) \\
  4)& \mathrm{~calculate~} w_t^k = \frac{p_\theta(s_t^k|s^{x^k_{t-1}}_{0:t-1}) p_\theta(o_t|s^k_{0:t})}{q_\phi(s^k_t|o_{1:t}, s^{x^k_{t-1}}_{0:t-1})}
\end{aligned}\right\} \mathrm{mutation}~~~~~~~~~~~~~~~~~~~~~~~~~~~~~~~~~~~~~~~~~
\end{equation*}
\vspace{-0.3cm}



until $K$ particles are updated. The posterior can be now approximated as $\sum_{k=1}^Kw^k_t\delta_{s_{0:t}^k}(s_{0:t})/\sum_{k=1}^K w^k_t$, where $\delta_{s_{0:t}^k}$ is a Dirac delta distribution concentrated at $s_{0:t}^k$.

Given observation $o$, a generator $p_\theta(s, o)$ and an inference approximate $q_\phi(s|o)$, the Monte Carlo objective \cite{Mnih2016MCObj, Maddison2017} with $K$ particles is a lower bound to $\log p_\theta(o)$ according to Jensen's inequality
\begin{align}
\log p_{\theta}(o) &\geq \mathbb{E}_{s\sim
q_{\phi}}\Big[ \log\frac{p_{\theta}(o|s)p_{\theta}(s)}{q_{\phi}(s|o)} \Big]\nonumber\\
&= \int\prod_{k=1}^K q_{\phi}(s^k|o) \log\frac{1}{K}\sum_{k=1}^K\frac{p_{\theta}(s^k, o)}{q_{\phi}(s^k|o)} \textnormal{d}s^{1:K}
\end{align}

Auto-encoding SMC \cite{Le2018AESMC} formalizes the ELBO objective for $\log p_\theta(o_{1:t})$ developed based on the estimator Eq.\eqref{eq:SMC_estimator} as
\begin{align}\label{eq:SMC_ELBO}
\mathrm{ELBO} &= \int\int\prod_{k=1}^K p_\theta(s_0^k)\prod_{\tau=1}^t\prod_{k=1}^K \frac{q_\phi(s_\tau^k|o_{1:\tau}, s_{0:\tau-1}^{x^k_{\tau-1}})w_{\tau-1}^{x^k_{\tau-1}}}{\sum_{k=1}^K w_{\tau-1}^k} \log \hat{p}_{\theta, \phi}^\mathrm{SMC} (o_{1:t}; x_{0:t-1}^{1:K}, s_{0:t}^{1:K})\mathrm{d}x_{0:t-1}^{1:K}\mathrm{d}s_{0:t}^{1:K}\nonumber\\
&= \mathbb{E}_{x_{0:t-1}^{1:K}, s_{0:t}^{1:K}\sim Q_{\mathrm{SMC}}}\Big[ \sum_{\tau=1}^t \log \big( \frac{1}{K}\sum_{k=1}^K w_\tau^k \big) \Big]
\end{align}
where $Q_{\mathrm{SMC}}$ is the distribution of SMC. The goal of model learning at time $t, ~t\geq 1$ is to maximize the lower bound to $\log p_\theta(o_{1:t})$, or minimize its negative
\begin{equation}\label{eq:SMC_ELBO_loss}
\mathcal{L}_\mathrm{ELBO}^{\theta, \phi} = - \frac{1}{t} \sum_{\tau=1}^t \log \big( \frac{1}{K}\sum_{k=1}^K w_\tau^k \big)
\end{equation}
If stationary state distribution $Q_{\mathrm{SMC}}$ is assumed, $\sum_{\tau=1}^t$ in Eq.\eqref{eq:SMC_ELBO} can be taken outside of the expectation, Eq.\eqref{eq:SMC_ELBO_loss} thus becomes an additive objective, allowing update using fractions of traces.

\subsection{Policy Gradient}
Under an MDP framework where environment state $s$ is fully observable, the goal of an RL agent is to maximize the expected long-term reward, or return $R_t= \sum_{\tau=t}\gamma^{\tau-t}r_\tau$, through optimizing a policy $\pi(a_t|s_t)$ in each state. State value under $\pi$ is defined as $V^\pi(s) = \mathbb{E}_\pi\big[ \sum_{\tau=t}\gamma^{\tau-t}r_\tau|s_t=s \big]$ and state-action value as $Q^\pi(s, a) = \mathbb{E}_\pi\big[ \sum_{\tau=t}\gamma^{\tau-t}r_\tau|s_t=s, a_t=a \big]$. The subscript $\pi$ denotes the action at each step dictated by $a_t\sim\pi(\cdot|s_t)$, and the stationary distribution over states implied accordingly.

Instead of learning $Q^\pi(s, a)$ which requires sufficient visits of all $a\in\mathcal{A}$ in each $s$, policy gradient methods explicitly represent $\pi$ with parameters $\omega$, and update them in the direction of $\mathbb{E}_\pi\big[ \nabla_\omega V^\pi(s_0) \big] = \mathbb{E}_\pi\big[ \sum_{t=0}\gamma^t\nabla_\omega\log\pi(a_t|s_t, \omega_t) R_t \big] = \mathbb{E}_\pi\big[ \sum_{t=0}\gamma^t\nabla_\omega\log\pi(a_t|s_t, \omega_t) Q^\pi(s_t, a_t) \big]$. Whereas $R_t$ based gradient estimator introduces lower bias but higher variance, function approximation based estimator incurs lower variance but higher bias \cite{Wang16sampleeff}. The two can be combined for bias-variance trade-off by subtracting a state-dependent but action-independent baseline $v_t$ from the state-action value estimate $q_t$ s.t. $\mathbb{E}_\pi\big[ q_t|s_t, a_t \big] = Q^\pi(s_t, a_t)$, to assuage variance, and calculating $q_t$ from reward or trace of rewards, i.e., bootstrapping, to reduce bias.

Classically the baseline $v_t$ is chosen as a function approximator $V_\kappa(s_t)$ that approximates $V^\pi(s_t)$ \cite{9Sutton98a, Wang16sampleeff, mniha16a3c, Espeholt2018IMPALA}. In this scenario, the difference is an estimate of the advantage function $A^\pi(s_t, a_t) = Q^\pi(s_t, a_t) - V^\pi(s_t)$. The asynchronous advantage actor-critic (A3C) \cite{mniha16a3c} proposes an on-policy estimate of the advantage function from a $k$-step forward-view temporal difference (TD) residual
\begin{equation}\label{eq:advantage_est_a3c}
\hat{A}^{\mathrm{A3C}}(s_t, a_t) = \sum_{i=0}^{k-1}\gamma^ir_{t+i} + \gamma^kV_\kappa(s_{t+k})-V_\kappa(s_t)
\end{equation}
The policy gradient at time $t$ can subsequently be calculated as $\nabla_\omega\log\pi(a_t|s_t, \omega_t)\hat{A}^{\mathrm{A3C}}(s_t, a_t)$.

\section{Auto-Encoding Heuristic Search}
This section introduces our method of sequential environment learning and heuristics in online planning. Pseudocode is provided in the supplementary material.

\subsection{Belief Update}
Generalization over belief states is more appropriate than over history trajectories as different histories may well result in more similar beliefs.
Belief at each step is determined by the transition and observation functions of the environment from predecessor belief, latest action and current observation
\begin{equation*}
b_t = \frac{p(o_t|s_t, a_{t-1}) \int p(s_t|s_{t-1}, a_{t-1})b_{t-1}\mathrm{d}s_{t-1}}{\int p(o_t|s_t, a_{t-1}) \int p(s_t|s_{t-1}, a_{t-1})b_{t-1}\mathrm{d}s_{t-1}\mathrm{d}s_t}
\end{equation*}

As with all posterior distributions, inference in non-trivial models require approximate approaches, among which SMC has shown remarkable performances \cite{Doucet2009SMC, Coquelin2009, Silver2010POMCP}. \cite{Igl2018} adapted SMC to learning belief state by incarnating the mutation step in SMC as an RNN, with another latent random variable $z_t$ that adds stochasticity
\begin{equation}
h_t = \Upsilon_\mathrm{RNN}(h_{t-1}, z_t, a_{t-1}, o_t|\theta)
\end{equation}
where $(z_t, a_{t-1}, o_t)$ is the input of the RNN at time $t$. The environment state is thereby the assembly of both latent variables $s_t = (h_t, z_t)$, associated with an importance weight $w_t^k$ that indicates how likely this particular sample is given historical trajectory.

Taking account of the RNN-boosted state representation, and the actions which are learned by the RL agent, the auto-encoding SMC becomes a generative model whose decoder, characterized by parameters $\theta$, consists of the transition function $p_\theta(z_t|h_{t-1}, a_{t-1})$, observation function $p_\theta(o_t|z_t, h_{t-1}, a_{t-1})$, initial density in $h$ space $p_\theta(h_0)$, and latent state transition $\Upsilon_\mathrm{RNN}(h_{t-1}, z_t, a_{t-1}, o_t|\theta)$; and whose encoder characterized by parameters $\phi$ $q_\phi(z_t|h_{t-1}, a_{t-1}, o_t)$ depends on history. The joint model likelihood conditional on actions $a_{0:t}$ is
\begin{align}\label{eq:model_joint_ll}
p_\theta (h_{0:t}, z_{1:t}, &o_{1:t}|a_{0:t}) = \nonumber\\
& p_\theta (h_0)\cdot
\prod_{\tau=1}^t p_\theta (z_\tau|h_{\tau-1}, a_{\tau-1})p_\theta (o_\tau|z_\tau, h_{\tau-1}, a_{\tau-1})
\delta_{\Upsilon_\mathrm{RNN}(h_{\tau-1}, z_\tau, a_{\tau-1}, o_\tau|\theta)}(h_\tau)
\end{align}
where $\delta_{\Upsilon_\mathrm{RNN}(h_{\tau-1}, z_\tau, a_{\tau-1}, o_\tau|\theta)}$ represents RNN state transition density. The belief state $b_t$ is a posterior distribution of the above model in the space of $(h_t, z_t)$.

To update a particle $s_t^k=(h_t^k, z_t^k)$, we first resample a predecessor from previous set $h_{t-1}^{x_{t-1}^k}$ according to relative importance weights. A $z_t^k$ is then sampled from $q_\phi (\cdot|h_{t-1}^{x_{t-1}^k}, a_{t-1}, o_t)$ using the reparameterization trick. Then we pass $(z_t^k, a_{t-1}, o_t)$ into $\Upsilon_\mathrm{RNN}$ whose last state is $h_{t-1}^{x_{t-1}^k}$ and get the new state $h_t^k$, and compute corresponding $w_t^k$. The procedure $b_t = \mathrm{UPDATE}(b_{t-1}, a_{t-1}, o_t)$ is encapsulated as follows:

1) resample index $x^k_{t-1}$ from predecessor set $x^k_{t-1}\sim w^{x^k_{t-1}}_{t-1}/\sum_{i=1}^K w^i_{t-1}$

2) sample latent variable $z^k_t\sim q_\phi(\cdot|h_{t-1}^{x^k_{t-1}}, a_{t-1}, o_t)$

3) update RNN state $h^k_t\gets\Upsilon_\mathrm{RNN}(h_{t-1}^{x^k_{t-1}}, z_t^k, a_{t-1}, o_t|\theta)$

4) calculate $w_t^k = \frac{p_\theta(z_t^k|h^{x^k_{t-1}}_{t-1}, a_{t-1}) p_\theta(o_t|z^k_t, h^{x^k_{t-1}}_{t-1}, a_{t-1})}{q_\phi(z^k_t|h_{t-1}^{x^k_{t-1}}, a_{t-1}, o_t)}$.

$\left\langle h^k_t, z^k_t, w^k_t \right\rangle^{1:K}$ is a particle assembly representation of $b_t$. To pass an argument to $V_\kappa$ or $\pi$ while maintaining uncertainty in $s_t$ after each transition, we need to aggregate information in the particle filter. The aggregation should be a permutation-invariant operation. Supported by the conviction from \cite{Murphy2019} that a permutation-sensitive function trained with permutation sampling retains tractability without notably compromising permutation invariability, we encode each $(h_t^k, z_t^k, w_t^k)$ through the same mechanism, and use a single permutation of the encoded features to learn the belief state. Specifically, each particle value $(h_t^k, z_t^k)$ and feature of weight $\varphi(w_t^k)$ are encoded into a belief feature with shared neural network parameters across particles, then the $K$ belief features are further encoded into a single belief state $b_t$. We will use the particle filter form $\mathcal{P}_t = \left\langle h^k_t, z^k_t, w^k_t \right\rangle^{1:K}$ and $b_t$ interchangeably as appropriate.

\subsection{Heuristics in Online Planning}
Policy gradient methods have stronger convergence guarantees than $Q$-learning variants \cite{9Sutton98a} since the target policy $\pi$ and value function $V_\kappa$ are updated smoothly via stochastic gradient descent with respect to their respective loss functions. However, the learning is potentially substantially off-policy as the target policy $\pi$ and behavior policy $\mu$ are probably disparate. In this circumstance, bootstrapping targets after taking into account the bias correction may account for scarce update information, detrimental to the asymptotic performance given limited patient trajectories. While we need to keep the target for $\pi$ bootstrapped based on previous discussion in subsection $3.2$, we compute the target for $V_\kappa$ from parallel heuristic search trees. Making additional use of the generative model, we evaluate each encountered belief state from model-reliant simulations to get an estimate closer to its true optimal value.

We learn a target policy via a policy gradient approach, actor-critic, representing the policy $\pi(a_t|b_t, \omega_t)$ (i.e. the \textit{actor}) and belief value function $V_\kappa(b_t)$ (i.e. the \textit{critic}) as function approximators. We will omit $\omega$ in $\pi$ to avoid clutter.

We exploit a set of parallel best-first suffix trees, both rooted at the current belief $b_t$, to evaluate it through simulations. We will omit subscript $t$ in search trees because each one of them involves only one actual belief state from environment. The root has a depth of $0$, denoted as $b^0$.

In each tree, we expand $N_s$ leaf nodes, starting from $b^0$. $b^d$ is expanded by a look-ahead tree policy $a_\mathcal{T}$ that exploits the decoder of the current auto-encoding model: first sample a subspace of the action space $\mathcal{A}$, and resample $n_z$ ancestor indices $x^i, ~i=1,\dots,n_z$ from $\mathcal{P}^d$ (i.e. selection step), then for each $a\in\mathcal{A}$ and each $x^i$, sample a $z^{d+1}_{a,i}$ according to the transition $p_\theta(\cdot|h^{d,x^i}, a)$, and for each $z^{d+1}_{a,i}$, simulate $n_{z,o} ~o^{d+1}_{a,i,j}$'s from $p_\theta(\cdot|z^{d+1}_{a,i}, h^{d,x^i}, a)$, $j=1,\dots,n_{z,o}$. Consequently, $n_z\cdot n_{z,o} ~o^{d+1}_{a,i,j}$'s will have been simulated for each $a$, each associated with a belief state $b^{d+1}_{a,i,j} = \mathrm{UPDATE}(b^d, a, o^{d+1}_{a,i,j})$ (i.e. mutation step). Then select immediate action that conforms with
\begin{equation*}
a_\mathcal{T}(b^d) = \mathop{\mathrm{arg\,max}}_{a \in\mathcal{A}} ~R(b^d, a)+ \frac{\gamma}{\eta^d(a)} \sum_{i=1}^{n_z}\sum_{j=1}^{n_{z,o}} \big( \frac{1}{K}\sum_{k=1}^K w^{d+1,k}_{a,i,j} \big) V_\kappa\big( b^{d+1}_{a,i,j} \big)
\end{equation*}
\begin{equation}\label{eq:look-ahead_action}
    =\mathop{\mathrm{arg\,max}}_{a \in\mathcal{A}}~ \Psi(b^d, a)~~~~~~~~~~~~~~~~~~~~~~~~~~~~~~~~~~~~~~~~~~~~
\end{equation}
where $R(b^d, a)$ is the reward by choosing $a$ in $b^d$, and $\eta^d(a) = \sum_{i=1}^{n_z}\sum_{j=1}^{n_{z,o}} \big( \frac{1}{K}\sum_{k=1}^K w^{d+1,k}_{a,i,j} \big)$ is a normalization constant for $a$.
Note that $\frac{1}{K}\sum_{k=1}^K w^{d+1,k}_{a,i,j} = \hat{p}_{\theta, \phi}(o^{d+1}_{a,i,j}|b^d, a)$ is an unbiased estimate Eq.\eqref{eq:SMC_estimator} of the marginal likelihood starting from parent $b^d$ conditioned on action $a$, assuming stationary distributions.
Therefore, if all functions were perfectly represented and approximating the exhaustion over possible successor observations by $n_z\cdot n_{z,o}$ samples of $o^{d+1}_{a,i,j}$, this action would become the true optimal policy for POMDPs \cite{52Ross2007AAO}
\begin{equation}\label{eq:optimal_action}
\pi^*(b_t) = \mathop{\mathrm{arg\,max}}_{a \in\mathcal{A}}
R(b_t, a) + \gamma\sum_{o_{t+1}}p(o_{t+1}|b_t, a)V^*\big( \tau(b_t, a, o_{t+1}) \big)
\end{equation}
where $V^*$ is the true optimal value function under $\pi^*$, and $\tau$ is the belief transition function commensurate with our $b^{d+1}_{a,i,j} = \mathrm{UPDATE}(b^d, a, o^{d+1}_{a,i,j})$.



$\Psi(b^d, a)$ can be thought of as a heuristic action value function that favors high-value weighted-average TD residuals with respect to the generative model. However, $\Psi$ only dictates immediate action to generate simulated experience rather than determine $\pi$ through maximization, which is known to suffer from stability and convergence issues \cite{Maei2009}. To capture potential multi-modeness and encourage exploration, we convert the policy in Eq.\eqref{eq:look-ahead_action} in density representation with softmax
\begin{equation}\label{eq:look-ahead_policy}
\hat{\pi}\big( a_\mathcal{T}(b^d)|b^d \big) =\frac{\exp\big( \Psi(b^d, a_\mathcal{T}(b^d))/\beta \big)}{Z(b^d)}
\end{equation}
where $\beta$ is a temperature parameter, $Z(b^d) = \int_\mathcal{A} \exp\big( \Psi(b^d, a)/\beta \big)\mathrm{d}a$ is a partition function. After choosing $a_\mathcal{T}(b^d)$, all sampled actions other than the selected one and their children are discarded.


Expansion of $b^d$ will therefore add $n_z\cdot n_{z, o}$ leaf nodes to the tree. We aim to encourage efficient expansions so that the tree is an any-time exploration mechanism. To this end, we propose expanding the node that contributes the largest payout to the root among all the leaf nodes $\mathcal{F}$: the node whose value is most likely to be back-propagated to that of the root
\begin{equation}\label{eq:best_leaf_node}
b^* = \mathop{\mathrm{arg\,max}}_{b \in\mathcal{F}}~\gamma^{D(b)} \prod_{\substack{d=1\\
b^0\rightarrow b}}^{D(b)} \frac{1}{\eta^{d-1}(*)}\big( \frac{1}{K}\sum_{k=1}^K w^{d,k} \big)
\end{equation}
where $D(b)$ is the depth of $b$, $*$ is a shorthand for the selected action, all weights and selected actions in question are from nodes in the path from root $b^0$ to $b$.

Once all the $N_s$ expansions have been investigated in silico, belief value estimates in the tree $v_\mathcal{T}$ are back-propagated in a bottom-up fashion to the root via Bellman update \cite{9Sutton98a, 52Ross2007AAO}
\begin{equation}
v_\mathcal{T}(b^d) = R\big(b^d, * \big) + \frac{\gamma}{\eta^d(*)} \sum_{i=1}^{n_z}\sum_{j=1}^{n_{z,o}} \big( \frac{1}{K}\sum_{k=1}^K w^{d+1,k}_{*,i,j} \big) V_\kappa\big( b^{d+1}_{*,i,j} \big)
\end{equation}
$\forall$ expanded $b^d$. Each node then has a value backed up as the weighted sum of one-step TD residuals from its direct children. The value estimates for the leaf nodes are provided by $V_{\kappa}$. $v_\mathcal{T}(b_t) = v_\mathcal{T}(b^0)$ is then the value estimated for $b_t$ from one search tree. We take the average across $N_e$ parallel trees $\sum_n^{N_e} v^n_\mathcal{T}(b_t)/N_e$ as the target for $V_{\kappa}(b_t)$, with the error loss being the mean-squared error between the two. If the model is exact, every application of the Bellman update makes the value estimation closer to the true optimal value \cite{Smith2004anal}. As a result, the value estimate for the root through value back-ups $v_\mathcal{T}(b_t)$ is closer to $V^*(b_t)$ than $V_{\kappa}(b_t)$ is.

Gradients are only propagated from environmental interactions, not from within simulations. Therefore, the trees only provide target for $V_\kappa$. Note that the target for $V_\kappa$ is decorrelated from $\pi$, and is instead always updated towards $V^*$ with respect to current environment model. In on-policy case, to make $\pi$ reflect $V_\kappa$, the gradients of $\pi$ should be expressed by reward traces generated by $\pi$, such as Eq.\eqref{eq:advantage_est_a3c}. In off-policy case, in contrast, since the reward traces are generated by $\mu$, both the calculation and the back-propagation of the TD error $r_s+\gamma V_\kappa(b_{s+1})-V_\kappa(b_s)$ at time $s$ for estimating the advantage function at time $t$ should be corrected by the difference in choosing $a_s$ in the two policies, or the per-decision importance sampling (IS) ratio $\frac{\pi(a_s|b_s)}{\mu(a_s|b_s)}$. To this end, analogously to the V-trace target introduced in \cite{Espeholt2018IMPALA}, we estimate the \textit{off-policy} advantage function as
\begin{equation}\label{eq:advantage_est_k}
\hat{A}^k(b_t, a_t) = \sum_{s=t}^{t+k-1}\gamma^{s-t} \Big( \prod_{i=t}^{s-1}c_i \Big) \delta_s
\end{equation}
where $\delta_s = \rho_s \big( r_s+\gamma V_\kappa(b_{s+1})-V_\kappa(b_s) \big)$ is the TD error for $V_\kappa$ at time $s$ weighed by the truncated IS ratio $\rho_s = \min\big( \frac{\pi(a_s|b_s)}{\mu(a_s|b_s)}, \rho^- \big)$, and $c_i = \lambda\min (\frac{\pi(a_s|b_s)}{\mu(a_s|b_s)}, c^-)$ is a further contracted IS ratio as is in Retrace\footnote{Define $\prod_{i=t}^{s-1}c_i=1$ when $t=s$.} \cite{Munos2016retrace}. The contractions effectively mitigate variance in the IS ratios, for details of which we refer readers to \cite{Munos2016retrace, Espeholt2018IMPALA}. The IS ratio $\rho_s$ within the definition of $\delta_s$ corrects bias inflicted by off-policyness, while $\prod_{i=t}^{s-1}c_i$ further reduces the impact of bias at time $s$ on the advantage estimate at $t$. This is similar to the generalized advantage estimator proposed in \cite{schulman2016}, the distinction being that we correct bias by coefficient $c_i$ instead of a constant $\lambda$. 

The superscript $k$ in Eq.\eqref{eq:advantage_est_k} denotes the estimate is calculated from an up-to-$k$ step trace. However, as we frame the medical application as a sparse-reward situation (details in the supplementary material), only the transitions to terminals being linked to non-zero rewards, we back-propagate these rewards to all previous steps in each respective trajectory. To this end, we estimate the advantage function with infinite horizon
\begin{equation}\label{eq:advantage_est_inf}
\hat{A}^\infty(b_t, a_t) = \sum_{s=t}^\infty\gamma^{s-t} \Big( \prod_{i=t}^{s-1}c_i \Big) \delta_s
\end{equation}
with a slight misusage of $\infty$ to denote the whole trajectory from $t$ onward till the terminal\footnote{We do not guarantee computation efficiency and storage with gradients propagated through full trajectory for longer episodes. As we are dealing with critical care data, patient trajectories are relatively short.}. As such, the maximum propagation length is the length of each admission trajectory.

\subsection{Learning Objective}
This subsection summarizes loss functions for training parameters jointly, with derivations spread over previous subsections. Admission trajectories are concatenated together. The mini batch size is denoted as $L$.

The loss introduced by model learning is
\begin{equation}
\mathcal{L}_{\mathrm{ELBO}}^{\theta, \phi} =  -\frac{1}{L}\sum_{\tau=t}^{t+L-1}\log \big( \frac{1}{K}\sum_{k=1}^K w_\tau^k \big)
\end{equation}

The policy loss $\mathcal{L}^{\theta,\phi,\omega}_{\pi}$ and the value function loss $\mathcal{L}^{\theta,\phi,\kappa}_{V}$ from RL online planning are
\begin{equation}
\mathcal{L}^{\theta,\phi,\omega}_{\pi} = -\frac{1}{L}\sum_{\tau=t}^{t+L-1}\log\pi(a_\tau|b_\tau, \omega_\tau)\hat{A}^{\infty}(b_\tau, a_\tau)
\end{equation}
\begin{equation}
\mathcal{L}^{\theta,\phi,\kappa}_{V} = \frac{1}{L}\sum_{\tau=t}^{t+L-1}\big( \frac{1}{N_e}\sum_n^{N_e} v^n_\mathcal{T}(b_\tau)-V_\kappa(b_\tau) \big)^2
\end{equation}
A policy entropy bonus could be imposed to mitigate premature policy convergence like in A3C
\begin{equation}
\mathcal{L}^{\theta,\phi,\omega}_{H} = -\frac{1}{L}\sum_{\tau=t}^{t+L-1}\int\pi(a|b_\tau)\log\pi(a|b_\tau)\mathrm{d}a
\end{equation}

Moreover, we estimate the behavior policy $\mu$ also from a neural network, parameterized by $\zeta$, the error loss being the negative loss probability of the actual behavior
\begin{equation}
\mathcal{L}^{\theta,\phi,\zeta}_{\mu} = -\frac{1}{L}\sum_{\tau=t}^{t+L-1} \log \mu(a_\tau|b_\tau, \zeta_\tau)
\end{equation}

The overall loss at update time $t$ is the sum of $\mathcal{L}_\mathrm{ELBO}^{\theta, \phi}$, $\mathcal{L}^{\theta,\phi,\omega}_\pi$, $\mathcal{L}^{\theta,\phi,\kappa}_V$ , $\mathcal{L}^{\theta,\phi,\omega}_H$ and $\mathcal{L}^{\theta,\phi,\zeta}_{\mu}$ rescaled by relative weights, which are hyperparameters. Note the overall loss is only computed from one thread.

\section{Experiments}
We test our method on a retrospective critical care dataset, MIMIC-III, where we treat each admission as an episode. Please see the supplementary material for more information on the cohort. One of our baselines is clinicians' behavior policy $\mu$, which we evaluate on-policy as the average empirical return of the test set. Further, to single out the advantage of involving heuristics in online planning, we compare AEHS to AE+A2C, our sequential generative model for POMDP with original synchronous A3C, or A2C, parallel workers. For AE+A2C we split all training steps (roughly) evenly into $N_e$ groups without breaking episodes. We also compare to when the patient-clinician interactions are modelled as MDPs by simplifying our generative model accordingly and interpolating missing data.

We use RMSProp \cite{Tieleman2012rmsprop} as the optimizer with shared statistics across threads. The discount factor in RL is $\gamma=0.99$. We set the population size of the particle filter $K=15$ and the number of threads $N_e=16$ for both AEHS and AE+A2C, the times of expansions $N_s=20$ and mini batch size $L=10$ for AEHS, and the temporal component of batch size for AE+A2C also $L=10$, leading to a total batch size of $L\times N_e$. In both contestants the TD errors are back-propagated over full trajectories. Note this is different in the original A2C implementation where the bootstrapped length is upper bounded by the number of steps run ahead $L$. This way one could maximally single out the effect of heuristic search. Please see the supplementary material for implementation details.

Learning curve visualizations are averaged over $10$ training instances, each consisting of $5$ epochs of random shuffles of all patient admissions in the training set. Shaded areas represent standard deviations.

The evaluation of our learned policy $\pi(a_t|b_t, \omega)$ using sequences generated by $\mu(a_t|b_t)$ is an instance of off-policy policy evaluation (OPPE). We implement weighted importance sampling (WIS) \cite{9Sutton98a} as a means to correct discrepancies between the probabilities of a trajectory under the two policies considering episodic tasks. The per-trajectory IS weight is defined as $W_n = \prod_{t=0}^{T_n-1}\frac{\pi(a_t|b_t, \omega)}{\mu(a_t|b_t)}$, the WIS return of a dataset is estimated as $R^\mathrm{WIS} = \frac{\sum_n^N W_nR_n}{\sum_n^N W_n}$, where $R_n$ is the empirical return of trajectory $n$. IS weights $W_n$ notoriously suffer from high variance, especially when $\pi$ is significantly dissimilar to $\mu$. This is detrimental, since trajectories corresponding to extremely small weights would effectively be ignored in evaluating the whole dataset.

\begin{figure}
    \centering{
        \begin{tabular}{ccc}
\includegraphics[width=0.31\textwidth]{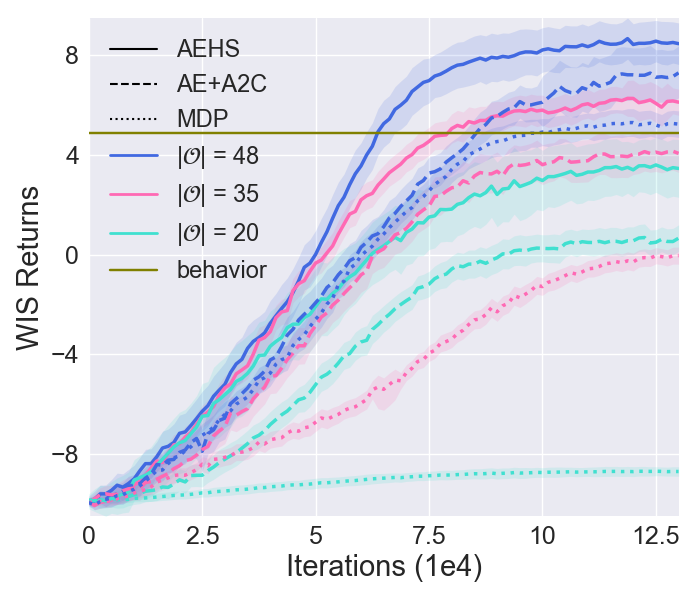}
            &
            \hspace*{-0.55cm}
\includegraphics[width=0.31\textwidth]{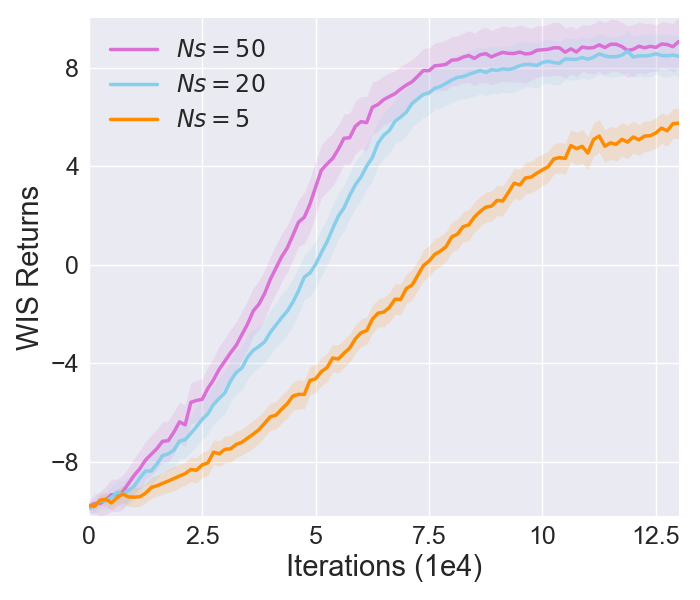}
            &
            \hspace*{-0.5cm}
\includegraphics[width=0.4\textwidth]{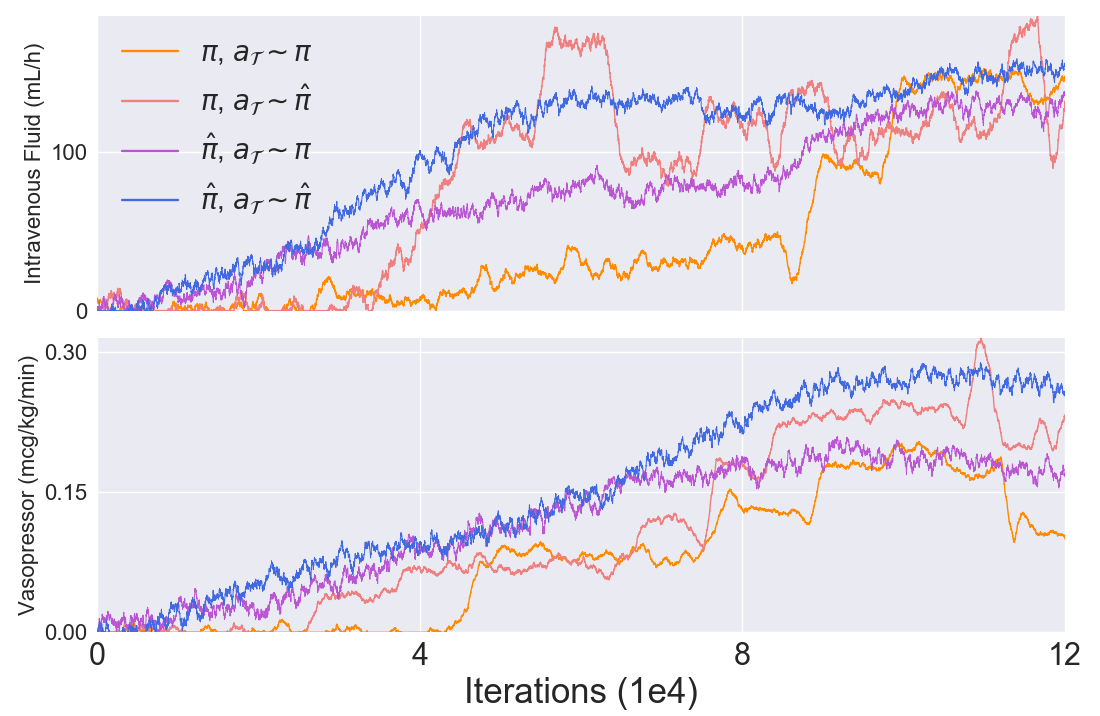}
\vspace{-0.2cm}\\
            ~~~~\scriptsize(a) & \hspace*{-0.5cm}~~~~~\scriptsize(b) & \hspace*{-0.5cm} ~~~~\scriptsize(c) \\
        \end{tabular}
    }
\vspace{-0.2cm}
\caption{(a) Validations on the test set, with parameters trained from different numbers of observed patient variables as an indicator of partial observability, and from corresponding sets of patient variables while assuming full observability. Results shown are the OPPE outcome of current parameters every training iteration. (b) Influence of the number of expansions in each tree search. (c) Action updates in a single belief, when actions in search trees are dictated by $\pi$ or $\hat{\pi}$. Negative actions during test are truncated at $0$.}
\label{fig:miscellaneous}
\end{figure}

Some works \cite{Hanna2017BootstrappingWM, Liu2018} have been dedicated to stable OPPE approaches. As the method for policy optimization is orthogonal to that for policy evaluation, and our major investigation is in the former, we simply clip each $W_n$ into $[W_-, W^-]$ as in \cite{Jagannatha2018}. We show with bootstrapping estimates in the supplementary material confidence bounds of OPPE, that with higher truncation levels, the variance in WIS returns is lower, but the policy we end up evaluating deviates more from the actual $\pi$ and more towards $\mu$.

\subsection{Overall Performance}
To see our robustness to data obscurity as well as to justify the incorporation of heuristics, we choose from the set of $48$ patient variables $35$ and then from them, $20$ patient variables as observations. For all of the three sets of observations, we first model the trajectories as POMDPs and implement AEHS and AE+A2C respectively. Missing data are \textit{not} interpolated for POMDPs. Then for the same three sets of observations, we use MDP model and apply to them the AEHS substituting the auto-encoding SMC with a conditional variational auto-encoder $p_\theta(o_0)$, $p_\theta(o_t|z_t, o_{t-1}, a_{t-1})$ and $q_\phi(z_t|o_t, o_{t-1}, a_{t-1})$, that directly maps $o_{t-1}, a_{t-1}$ into $o_t$. Missing data are linearly interpolated in MDPs\footnote{Binary variables in MDPs are interpolated with sample-and-hold.}. After each training iteration (i.e. $L$ steps), the model and RL agent is validated on the test set, with each per-trajectory IS weight clipped into $[1e-30, 1e4]$.

Results in Figure \ref{fig:miscellaneous} $(a)$ show that, modelling as MDPs is more susceptible to the number of patient variables involved and consistently yields poorer performances despite the observation dimensionality than modelling as POMDPs, vindicating that patient-clinician interactions are de facto POMDPs. Moreover, as the incompleteness of observations increases in POMDP framework, both AEHS and AE+A2C learn worse, but AEHS deteriorates slightly less than AE+A2C, and learns faster and better across all three cases. The horizontal line manifests the on-policy policy evaluation result of clinicians' behavior policy. We note that, given limited trajectories, upon convergence AE+A2C ceases to surpass clinical baseline at $35$ patient variables, while AEHS ceases to do so at $20$ patient variables. In the supplementary material we show that with $48$ patient variables, the $95 \%$ confidence lower bound on the WIS return of the test set under this specific truncation level of IS weights is $7.030$, while the average empirical return per trajectory under $\mu$ is $4.867$.

\subsection{Efficiency of Heuristics}
To test whether our heuristics are applied efficiently, we first investigate the influence of the number of node expansions $N_s$ in each suffix tree. As is shown in Figure \ref{fig:miscellaneous} $(b)$, increasing $N_s$ does not dramatically boost performance. This is because the trees are best-first: Eq.\eqref{eq:best_leaf_node} always selects the most influential node to expand, implying that there is no need to expand many more nodes to evaluate the root.

We then track the action values, dosages of vasopressor and of intravenous fluid, in a single belief of a single training instance. Specifically, we compare how the target policy $\pi$ is updated and how the look-ahead policy $\hat{\pi}$ would appear when selecting actions $a_{\mathcal{T}}$ by $\pi$ or $\hat{\pi}$ respectively during tree explorations, as real actions are unanimously dictated by $\mu$. The results visualized in Figure \ref{fig:miscellaneous} $(c)$ suggest that, $\hat{\pi}$ tends to update ahead of $\pi$ like a precursor, and updates more stably on the whole. In addition, $\pi$ is able to update faster when simulations are dictated by $\hat{\pi}$ than by $\pi$. This is because $\hat{\pi}$ accounts for more information for planning than $\pi$ by probing into the values of multiple reachable belief states. Expanding the suffix trees with $\hat{\pi}$ therefore helps with the efficiency of finding a target for $V_\kappa$ that is closer to $V^*$.

\section{Conclusion}
We propose a novel POMDP approach to optimizing sepsis treatment from a retrospective critical care database. To encourage belief inference, we build an auto-encoding SMC and incorporate the ELBO as an auxiliary loss. Secondly, we propose a heuristic planning architecture that exploits best-first suffix trees to improve data efficiency. Thirdly, lagged learning effect induced from reward sparsity is mitigated by back-propagating over full trajectories, alongside an off-policy estimate of policy gradient is devised with bias correction and variance control. Our experimental results demonstrate that, 1) POMDP is indeed a more appropriate model for patient-clinician interactions than MDP, 2) our approach is able to surpass clinicians' baseline by a remarkable margin with high probability in OPPE terms, and 3) our heuristics in tree policy and nodes to expand consistently improves planning performance and data efficiency.

\bibliographystyle{sort}
\bibliographystyle{icml2019}

\begin{thebibliography}{63}
\providecommand{\natexlab}[1]{#1}
\providecommand{\url}[1]{\texttt{#1}}
\expandafter\ifx\csname urlstyle\endcsname\relax
  \providecommand{\doi}[1]{doi: #1}\else
  \providecommand{\doi}{doi: \begingroup \urlstyle{rm}\Url}\fi

\bibitem[Asoh et~al.(2013)Asoh, Shiro, Akaho, Kamishima, Hasida, Aramaki, and
  Kohro]{19Asoh2013AAI}
Hideki Asoh, Masanori Shiro, Shotaro Akaho, Toshihiro Kamishima, Koiti Hasida,
  Eiji Aramaki, and Takahide Kohro.
\newblock An application of inverse reinforcement learning to medical records
  of diabetes treatment.
\newblock In \emph{European Conference on Machine Learning and Principles and
  Practice of Knowledge Discovery in Databases}, Sep 2013.

\bibitem[Azizzadenesheli et~al.(2016)Azizzadenesheli, Lazaric, and
  Anandkumar]{Azizzadenesheli2016}
Kamyar Azizzadenesheli, Alessandro Lazaric, and Animashree Anandkumar.
\newblock Reinforcement learning of pomdps using spectral methods.
\newblock In \emph{29th Annual Conference on Learning Theory}, volume~49 of
  \emph{Proceedings of Machine Learning Research}, pages 193--256, Columbia
  University, New York, New York, USA, 23--26 Jun 2016. PMLR.

\bibitem[Bakker(2001)]{Bakker2001}
Bram Bakker.
\newblock Reinforcement learning with long short-term memory.
\newblock In \emph{Proceedings of the 14th International Conference on Neural
  Information Processing Systems: Natural and Synthetic}, 2001.

\bibitem[Barth{-}Maron et~al.(2018)Barth{-}Maron, W.~Hoffman, Budden, Dabney,
  Horgan, Tirumala, Muldal, Heess, and Lillicrap]{Barth2018}
Gabriel Barth{-}Maron, Matthew W.~Hoffman, David Budden, Will Dabney, Dan
  Horgan, Dhruva Tirumala, Alistair Muldal, Nicolas Heess, and Timothy~P.
  Lillicrap.
\newblock Distributed distributional deterministic policy gradients.
\newblock \emph{ICLR}, abs/1804.08617, 2018.

\bibitem[Bothe et~al.(2013)Bothe, Dickens, Reichel, Tellmann, Ellger, Westphal,
  and Faisal]{11Bothe2013TheUO}
Melanie Bothe, Luke W.~F. Dickens, Katrin Reichel, Arn Tellmann, Bjoern Ellger,
  Martin Westphal, and Aldo~A. Faisal.
\newblock The use of reinforcement learning algorithms to meet the challenges
  of an artificial pancreas.
\newblock \emph{Expert review of medical devices}, 10\penalty0 (5):\penalty0
  661--73, 2013.

\bibitem[Brown et~al.(2013)Brown, Lanspa, Jones, Kuttler, Li, Carlson, Miller,
  Hirshberg, Grissom, and Morris]{Brown2013vasopressor}
Samuel~M. Brown, Michael~J. Lanspa, Jason~P. Jones, Kathryn~G. Kuttler, Yao Li,
  Rick Carlson, Russell~R. Miller, Eliotte~L. Hirshberg, Colin~K. Grissom, and
  Alan~H. Morris.
\newblock Survival after shock requiring high-dose vasopressor therapy.
\newblock \emph{Chest}, 143\penalty0 (3):\penalty0 664 -- 671, 2013.

\bibitem[Cassandra et~al.(1997)Cassandra, Littman, and
  Zhang]{38Cassandra97incrementalpruning:}
Anthony Cassandra, Michael~L. Littman, and Nevin~L. Zhang.
\newblock Incremental pruning: A simple, fast, exact method for partially
  observable markov decision processes.
\newblock In \emph{In Proceedings of the Thirteenth Conference on Uncertainty
  in Artificial Intelligence}, pages 54--61. Morgan Kaufmann Publishers, 1997.

\bibitem[Cassandra et~al.(1994)Cassandra, Kaelbling, and
  Littman]{35Cassandra1994AOI}
Anthony~R. Cassandra, Leslie~P. Kaelbling, and Michael~L. Littman.
\newblock Acting optimally in partially observable stochastic domains.
\newblock \emph{Twelfth National Conference on Artificial Intelligence
  (AAAI-94)}, pages 1023--1028, 1994.

\bibitem[Coquelin et~al.(2009)Coquelin, Deguest, and Munos]{Coquelin2009}
Pierre-arnaud Coquelin, Romain Deguest, and R\'{e}mi Munos.
\newblock Particle filter-based policy gradient in pomdps.
\newblock In D.~Koller, D.~Schuurmans, Y.~Bengio, and L.~Bottou, editors,
  \emph{Advances in Neural Information Processing Systems 21}, pages 337--344.
  2009.

\bibitem[Doshi-velez(2009)]{47Doshi-velez2009TIPO}
Finale Doshi-velez.
\newblock The infinite partially observable markov decision process.
\newblock In \emph{Advances in Neural Information Processing Systems 22}, pages
  477--485. 2009.

\bibitem[Doshi-Velez et~al.(2012)Doshi-Velez, Pineau, and
  Roy]{49DOSHIVELEZ2012RLWLR}
Finale Doshi-Velez, Joelle Pineau, and Nicholas Roy.
\newblock Reinforcement learning with limited reinforcement: Using bayes risk
  for active learning in pomdps.
\newblock \emph{Artificial Intelligence}, 187-188:\penalty0 115 -- 132, 2012.

\bibitem[Doshi-Velez et~al.(2015)Doshi-Velez, Pfau, Wood, and
  Roy]{DoshiVelez2015}
Finale Doshi-Velez, David Pfau, Frank Wood, and Nicholas Roy.
\newblock Bayesian nonparametric methods for partially-observable reinforcement
  learning.
\newblock \emph{IEEE Transactions on Pattern Analysis and Machine
  Intelligence}, 37\penalty0 (2):\penalty0 394--407, 2015.

\bibitem[Doucet and Johansen(2009)]{Doucet2009SMC}
Arnaud Doucet and Adam~M. Johansen.
\newblock A tutorial on particle filtering and smoothing: Fifteen years later.
\newblock In \emph{in Oxford Handbook of Nonlinear Filtering}. University
  Press, 2009.

\bibitem[Ernst et~al.(2006)Ernst, Stan, Goncalves, and
  Wehenkel]{10Ernst2006CDB}
Damien Ernst, Guy-Bart Stan, Jorge Goncalves, and Louis Wehenkel.
\newblock Clinical data based optimal sti strategies for hiv: a reinforcement
  learning approach.
\newblock In \emph{Proceedings of the 45th IEEE Conference on Decision and
  Control}, pages 667--672, Dec 2006.

\bibitem[Espeholt et~al.(2018)Espeholt, Soyer, Munos, Simonyan, Mnih, Ward,
  Doron, Firoiu, Harley, Dunning, Legg, and Kavukcuoglu]{Espeholt2018IMPALA}
Lasse Espeholt, Hubert Soyer, Remi Munos, Karen Simonyan, Vlad Mnih, Tom Ward,
  Yotam Doron, Vlad Firoiu, Tim Harley, Iain Dunning, Shane Legg, and Koray
  Kavukcuoglu.
\newblock {IMPALA}: Scalable distributed deep-{RL} with importance weighted
  actor-learner architectures.
\newblock In \emph{Proceedings of the 35th International Conference on Machine
  Learning}, volume~80 of \emph{Proceedings of Machine Learning Research},
  pages 1407--1416, Stockholmsmässan, Stockholm Sweden, 10--15 Jul 2018. PMLR.

\bibitem[Esteva et~al.(2017)Esteva, Kuprel, Novoa, Ko, Swetter, Blau, and
  Thrun]{esteva2017dermatologist}
Andre Esteva, Brett Kuprel, Roberto~A. Novoa, Justin Ko, Susan~M. Swetter,
  Helen~M. Blau, and Sebastian Thrun.
\newblock Dermatologist-level classification of skin cancer with deep neural
  networks.
\newblock \emph{Nature}, 542\penalty0 (7639):\penalty0 115, 2017.

\bibitem[Gulshan et~al.(2016)Gulshan, Peng, Coram, Stumpe, Wu, Narayanaswamy,
  Venugopalan, Widner, Madams, Cuadros, Kim, Raman, Nelson, Mega, and
  Webster]{gulshan2016development}
Varun Gulshan, Lily Peng, Marc Coram, Martin~C. Stumpe, Derek Wu, Arunachalam
  Narayanaswamy, Subhashini Venugopalan, Kasumi Widner, Tom Madams, Jorge
  Cuadros, Ramasamy Kim, Rajiv Raman, Philip~C. Nelson, Jessica~L. Mega, and
  Dale~R. Webster.
\newblock Development and validation of a deep learning algorithm for detection
  of diabetic retinopathy in retinal fundus photographs.
\newblock \emph{Jama}, 316\penalty0 (22):\penalty0 2402--2410, 2016.

\bibitem[Hanna et~al.(2017)Hanna, Stone, and Niekum]{Hanna2017BootstrappingWM}
Josiah~P. Hanna, Peter Stone, and Scott Niekum.
\newblock Bootstrapping with models: Confidence intervals for off-policy
  evaluation.
\newblock In \emph{AAMAS}, 2017.

\bibitem[Hausknecht and Stone(2015)]{Hausknecht2015}
Matthew~J. Hausknecht and Peter Stone.
\newblock Deep recurrent q-learning for partially observable mdps.
\newblock \emph{arxiv.org/abs/1507.06527}, 2015.

\bibitem[Hauskrecht(2000)]{40Hauskrecht2000VAP}
Milos Hauskrecht.
\newblock Value-function approximations for partially observable markov
  decision processes.
\newblock \emph{J. Artif. Int. Res.}, 13\penalty0 (1):\penalty0 33--94, August
  2000.

\bibitem[Hauskrecht and Fraser(2000)]{Hauskrecht2000}
Milos Hauskrecht and Hamish S.~F. Fraser.
\newblock Planning treatment of ischemic heart disease with partially
  observable markov decision processes.
\newblock \emph{Artificial intelligence in medicine}, 18 3:\penalty0 221--44,
  2000.

\bibitem[Horgan et~al.(2018)Horgan, Quan, Budden, Barth-Maron, Hessel, van
  Hasselt, and Silver]{Horgan2018}
Dan Horgan, John Quan, David Budden, Gabriel Barth-Maron, Matteo Hessel, Hado
  van Hasselt, and David Silver.
\newblock Distributed prioritized experience replay.
\newblock \emph{arxiv.org/abs/1803.00933}, 2018.

\bibitem[Igl et~al.(2018)Igl, Zintgraf, Le, Wood, and Whiteson]{Igl2018}
Maximilian Igl, Luisa~M. Zintgraf, Tuan~Anh Le, Frank Wood, and Shimon
  Whiteson.
\newblock Deep variational reinforcement learning for pomdps.
\newblock In \emph{ICML}, 2018.

\bibitem[Jaderberg et~al.(2016)Jaderberg, Mnih, Czarnecki, Schaul, Leibo,
  Silver, and Kavukcuoglu]{Jaderberg2016}
Max Jaderberg, Volodymyr Mnih, Wojciech~Marian Czarnecki, Tom Schaul, Joel~Z.
  Leibo, David Silver, and Koray Kavukcuoglu.
\newblock Reinforcement learning with unsupervised auxiliary tasks.
\newblock \emph{CoRR}, abs/1611.05397, 2016.

\bibitem[Jagannatha et~al.(2018)Jagannatha, Thomas, and Yu]{Jagannatha2018}
Abhyuday Jagannatha, Philip Thomas, and Hone Yu.
\newblock Towards high confidence off-policy reinforcement learning for
  clinical applications.
\newblock \emph{CausalML Workshop, ICML}, 2018.

\bibitem[Johnson et~al.(2017)Johnson, Pollard, Shen, Lehman, Feng, Ghassemi,
  Moody, Szolovits, Anthony~Celi, and Mark]{Johnson2017MAF}
Alistair~E.W. Johnson, Tom~J. Pollard, Lu~Shen, Li-wei~H. Lehman, Mengling
  Feng, Mohammad Ghassemi, Benjamin Moody, Peter Szolovits, Leo Anthony~Celi,
  and Roger~G. Mark.
\newblock Mimic-iii, a freely accessible critical care database.
\newblock 3, 2017.

\bibitem[Kapturowski et~al.(2019)Kapturowski, Ostrovski, Quan, Munos, and
  Dabney]{Kapturowski2019}
Steven Kapturowski, Georg Ostrovski, John Quan, Remi Munos, and Will Dabney.
\newblock Recurrent experience replay in distributed reinforcement learning.
\newblock \emph{ICLR}, 2019.

\bibitem[Komorowski et~al.(2018)Komorowski, Celi, Badawi, Gordon, and
  Faisal]{Komorowski2018naturemed}
Matthieu Komorowski, Leo~A. Celi, Omar Badawi, Anthony~C. Gordon, and A.~Aldo
  Faisal.
\newblock The artificial intelligence clinician learns optimal treatment
  strategies for sepsis in intensive care.
\newblock In \emph{Nature Medicine}, volume~24, pages 1716--1720. 2018.

\bibitem[Kurniawati et~al.(2008)Kurniawati, Hsu, and Lee]{Kurniawati2008SARSOP}
Hanna Kurniawati, David Hsu, and Wee~Sun Lee.
\newblock {SARSOP}: Efficient point-based {POMDP} planning by approximating
  optimally reachable belief spaces.
\newblock In \emph{Proceedings of Robotics: Science and Systems IV}, Zurich,
  Switzerland, June 2008.

\bibitem[Le et~al.(2018)Le, Igl, Rainforth, Jin, and Wood]{Le2018AESMC}
Tuan~Anh Le, Maximilian Igl, Tom Rainforth, Tom Jin, and Frank Wood.
\newblock Auto-encoding sequential monte carlo.
\newblock In \emph{International Conference on Learning Representations}, 2018.

\bibitem[Litjens et~al.(2017)Litjens, Kooi, Bejnordi, Setio, Ciompi,
  Ghafoorian, van~der Laak, van Ginneken, and S{\'a}nchez]{litjens2017survey}
Geert Litjens, Thijs Kooi, Babak~E. Bejnordi, Arnaud A.~A. Setio, Francesco
  Ciompi, Mohsen Ghafoorian, Jeroen A. W.~M. van~der Laak, Bram van Ginneken,
  and Clara~I S{\'a}nchez.
\newblock A survey on deep learning in medical image analysis.
\newblock \emph{Medical image analysis}, 42:\penalty0 60--88, 2017.

\bibitem[Littman et~al.(1995)Littman, Cassandra, and
  Kaelbling]{33Littman1995LPP}
Michael~L. Littman, Anthony~R. Cassandra, and Leslie~Pack Kaelbling.
\newblock Learning policies for partially observable environments: Scaling up.
\newblock In \emph{Proceedings of the Twelfth International Conference on
  International Conference on Machine Learning}, pages 362--370, San Francisco,
  CA, USA, 1995. Morgan Kaufmann Publishers Inc.

\bibitem[Liu et~al.(2018)Liu, Gottesman, Raghu, Komorowski, Faisal,
  Doshi-Velez, and Brunskill]{Liu2018}
Yao Liu, Omer Gottesman, Aniruddh Raghu, Matthieu Komorowski, Aldo~A. Faisal,
  Finale Doshi-Velez, and Emma Brunskill.
\newblock Representation balancing mdps for off-policy policy evaluation.
\newblock In \emph{Advances in Neural Information Processing Systems 31}, pages
  2649--2658. 2018.

\bibitem[Lizotte and Laber(2016)]{25Lizotte2016MOMD}
Daniel~J. Lizotte and Eric~B. Laber.
\newblock Multi-objective markov decision processes for data-driven decision
  support.
\newblock \emph{Journal of Machine Learning Research}, 17\penalty0
  (211):\penalty0 1--28, 2016.

\bibitem[Lowery and Faisal(2013)]{Lowery2013}
Cristobal Lowery and Aldo~A. Faisal.
\newblock Towards efficient, personalized anesthesia using continuous
  reinforcement learning for propofol infusion control.
\newblock In \emph{2013 6th International IEEE/EMBS Conference on Neural
  Engineering (NER)}, pages 1414--1417, Nov 2013.
\newblock \doi{10.1109/NER.2013.6696208}.

\bibitem[Maddison et~al.(2017)Maddison, Lawson, Tucker, Heess, Norouzi, Mnih,
  Doucet, and Teh]{Maddison2017}
Chris~J. Maddison, John Lawson, George Tucker, Nicolas Heess, Mohammad Norouzi,
  Andriy Mnih, Arnaud Doucet, and Yee Teh.
\newblock Filtering variational objectives.
\newblock In \emph{Advances in Neural Information Processing Systems 30}, pages
  6573--6583. 2017.

\bibitem[Maei et~al.(2009)Maei, Szepesv\'{a}ri, Bhatnagar, Precup, Silver, and
  Sutton]{Maei2009}
Hamid~R. Maei, Csaba Szepesv\'{a}ri, Shalabh Bhatnagar, Doina Precup, David
  Silver, and Richard~S. Sutton.
\newblock Convergent temporal-difference learning with arbitrary smooth
  function approximation.
\newblock In \emph{Proceedings of the 22Nd International Conference on Neural
  Information Processing Systems}, pages 1204--1212, 2009.

\bibitem[Mnih and Rezende(2016)]{Mnih2016MCObj}
Andriy Mnih and Danilo~Jimenez Rezende.
\newblock Variational inference for monte carlo objectives.
\newblock \emph{CoRR}, abs/1602.06725, 2016.

\bibitem[Mnih et~al.(2015)Mnih, Kavukcuoglu, Silver, Rusu, Veness, Bellemare,
  Graves, Riedmiller, Fidjeland, Ostrovski, Petersen, Beattie, Sadik,
  Antonoglou, King, Kumaran, Wierstra, Legg, and Hassabis]{Mnih2015nature}
Volodymyr Mnih, Koray Kavukcuoglu, David Silver, Andrei~A. Rusu, Joel Veness,
  Marc~G. Bellemare, Alex Graves, Martin Riedmiller, Andreas~K. Fidjeland,
  Georg Ostrovski, Stig Petersen, Charles Beattie, Amir Sadik, Ioannis
  Antonoglou, Helen King, Dharshan Kumaran, Daan Wierstra, Shane Legg, and
  Demis Hassabis.
\newblock Human-level control through deep reinforcement learning.
\newblock \emph{Nature}, pages 518--529, 2015.

\bibitem[Mnih et~al.(2016)Mnih, Badia, Mirza, Graves, Lillicrap, Harley,
  Silver, and Kavukcuoglu]{mniha16a3c}
Volodymyr Mnih, Adria~Puigdomenech Badia, Mehdi Mirza, Alex Graves, Timothy
  Lillicrap, Tim Harley, David Silver, and Koray Kavukcuoglu.
\newblock Asynchronous methods for deep reinforcement learning.
\newblock In \emph{Proceedings of The 33rd International Conference on Machine
  Learning}, volume~48 of \emph{Proceedings of Machine Learning Research},
  pages 1928--1937. PMLR, 20--22 Jun 2016.

\bibitem[Munos et~al.(2016)Munos, Stepleton, Harutyunyan, and
  Bellemare]{Munos2016retrace}
R{\'{e}}mi Munos, Tom Stepleton, Anna Harutyunyan, and Marc~G. Bellemare.
\newblock Safe and efficient off-policy reinforcement learning.
\newblock \emph{CoRR}, abs/1606.02647, 2016.

\bibitem[Murphy et~al.(2019)Murphy, Srinivasan, Rao, and Ribeiro]{Murphy2019}
Ryan~L. Murphy, Balasubramaniam Srinivasan, Vinayak Rao, and Bruno Ribeiro.
\newblock Janossy pooling: Learning deep permutation-invariant functions for
  variable-size inputs.
\newblock \emph{ICLR}, 2019.

\bibitem[Nemati et~al.(2016)Nemati, Ghassemi, and
  Clifford]{24Nemati2016OptimalMD}
Shamim Nemati, Mohammad~M. Ghassemi, and Gari~D. Clifford.
\newblock Optimal medication dosing from suboptimal clinical examples: A deep
  reinforcement learning approach.
\newblock \emph{2016 38th Annual International Conference of the IEEE
  Engineering in Medicine and Biology Society (EMBC)}, pages 2978--2981, 2016.

\bibitem[Pineau et~al.(2003)Pineau, Gordon, and Thrun]{29Pineau2003PBVI}
Joelle Pineau, Geoffrey~J. Gordon, and Sebastian Thrun.
\newblock Point-based value iteration: An anytime algorithm for pomdps.
\newblock In \emph{IJCAI}, pages 1025--1032, 2003.

\bibitem[Prasad et~al.(2017)Prasad, Cheng, Chivers, Draugelis, and
  Engelhardt]{12Prasad2017MVI}
Niranjani Prasad, Li-Fang Cheng, Corey Chivers, Michael Draugelis, and
  Barbara~E Engelhardt.
\newblock A reinforcement learning approach to weaning of mechanical
  ventilation in intensive care units.
\newblock \emph{Proceedings of Uncertainty in Artificial Intelligence (UAI)},
  2017.

\bibitem[Ross and Chaib-Draa(2007)]{52Ross2007AAO}
St{\'e}phane Ross and Brahim Chaib-Draa.
\newblock Aems: An anytime online search algorithm for approximate policy
  refinement in large pomdps.
\newblock In \emph{Proceedings of the 20th International Joint Conference on
  Artifical Intelligence}, pages 2592--2598, San Francisco, CA, USA, 2007.
  Morgan Kaufmann Publishers Inc.

\bibitem[Ross et~al.(2007)Ross, Brahim, and Pineau]{Ross2007Bayespomdp}
Stephane Ross, Chaib-draa Brahim, and Joelle Pineau.
\newblock Bayes-adaptive pomdps.
\newblock In \emph{Advances in Neural Information Processing Systems 20}, pages
  1225--1232. 2007.

\bibitem[Schulman et~al.(2016)Schulman, Moritz, Levine, I.~Jordan, and
  Abbeel]{schulman2016}
John Schulman, Philipp Moritz, Sergey Levine, Michael I.~Jordan, and Pieter
  Abbeel.
\newblock High-dimensional continuous control using generalized advantage
  estimation.
\newblock \emph{ICLR}, abs/1506.02438, 2016.

\bibitem[Shortreed et~al.(2011)Shortreed, Laber, Lizotte, Stroup, Pineau, and
  Murphy]{8Shortreed2011ISCD}
Susan~M. Shortreed, Eric Laber, Daniel~J. Lizotte, T.~Scott Stroup, Joelle
  Pineau, and Susan~A. Murphy.
\newblock Informing sequential clinical decision-making through reinforcement
  learning: an empirical study.
\newblock \emph{Machine Learning}, 84\penalty0 (1):\penalty0 109--136, Jul
  2011.

\bibitem[Silver and Veness(2010)]{Silver2010POMCP}
David Silver and Joel Veness.
\newblock Monte-carlo planning in large pomdps.
\newblock In \emph{Advances in Neural Information Processing Systems 23}, pages
  2164--2172. 2010.

\bibitem[Singer et~al.(2016)Singer, Deutschman, Seymour, Shankar-Hari, Annane,
  Bauer, Bellomo, Bernard, Chiche, Coopersmith, Hotchkiss, Mitchell, Marshall,
  Martin, Opal, Rubenfeld, van~der Poll, Vincent, and Angus]{Singer2016}
Mervyn Singer, Clifford~S. Deutschman, Christopher~Warren Seymour, Manu
  Shankar-Hari, Djillali Annane, Michael Bauer, Rinaldo Bellomo, Gordon~R.
  Bernard, Jean-Daniel Chiche, Craig~M. Coopersmith, Richard~S. Hotchkiss, Levy
  Mitchell, John~C. Marshall, Greg~S. Martin, Steven~M. Opal, Gordon~D.
  Rubenfeld, Tom van~der Poll, Jean-Louis Vincent, and Derek~C. Angus.
\newblock The third international consensus definitions for sepsis and septic
  shock (sepsis-3).
\newblock \emph{JAMA}, 315\penalty0 (8):\penalty0 801--810, 2016.

\bibitem[Smith and Simmons(2004{\natexlab{a}})]{42Smith2004HSV}
Trey Smith and Reid Simmons.
\newblock Heuristic search value iteration for pomdps.
\newblock In \emph{Proceedings of the 20th Conference on Uncertainty in
  Artificial Intelligence}, pages 520--527, Arlington, Virginia, United States,
  2004{\natexlab{a}}. AUAI Press.

\bibitem[Smith and Simmons(2004{\natexlab{b}})]{Smith2004anal}
Trey Smith and Reid Simmons.
\newblock Point-based pomdp algorithms: imporved analysis and implementation.
\newblock \emph{arxiv.org/pdf/1207.1412}, 2004{\natexlab{b}}.

\bibitem[Sondik(1978)]{30Sondik1978TOCP}
Edward~J. Sondik.
\newblock The optimal control of partially observable markov processes over the
  infinite horizon: Discounted costs.
\newblock \emph{Operations Research}, 26\penalty0 (2):\penalty0 282--304, 1978.

\bibitem[Spaan and Vlassis(2004)]{39Spaan2004APPOMDP}
Matthijs T.~J. Spaan and Nikos Vlassis.
\newblock A point-based pomdp algorithm for robot planning.
\newblock In \emph{Proceedings. ICRA '04. 2004 IEEE International Conference on
  Robotics and Automation}, volume~3, pages 2399--2404, April 2004.

\bibitem[Sutton and Barto(1998)]{9Sutton98a}
Richard~S. Sutton and Andrew~G. Barto.
\newblock \emph{Reinforcement Learning : An Introduction}.
\newblock MIT Press, 1998.

\bibitem[Tieleman and Hinton(2012)]{Tieleman2012rmsprop}
Tijmen Tieleman and Geoffrey Hinton.
\newblock Lecture 6.5-rmsprop: Divide the gradient by a running average of its
  recent magnitude.
\newblock \emph{COURSERA: Neural Networks for Machine Learning}, 2012.

\bibitem[Trinh et~al.(2018)Trinh, Dai, Luong, and Le]{trinh18}
Trieu Trinh, Andrew Dai, Thang Luong, and Quoc Le.
\newblock Learning longer-term dependencies in {RNN}s with auxiliary losses.
\newblock In \emph{Proceedings of the 35th International Conference on Machine
  Learning}, volume~80, pages 4965--4974, 2018.

\bibitem[Tsoukalas et~al.(2015)Tsoukalas, Albertson, and
  Tagkopoulos]{Tsoukalas2015}
Athanasios Tsoukalas, Timothy Albertson, and Ilias Tagkopoulos.
\newblock From data to optimal decision making: A data-driven, probabilistic
  machine learning approach to decision support for patients with sepsis.
\newblock In \emph{JMIR medical informatics}, 2015.

\bibitem[Wang et~al.(2016)Wang, Bapst, Heess, Mnih, Munos, Kavukcuoglu, and
  de~Freitas]{Wang16sampleeff}
Ziyu Wang, Victor Bapst, Nicolas Heess, Volodymyr Mnih, R{\'{e}}mi Munos, Koray
  Kavukcuoglu, and Nando de~Freitas.
\newblock Sample efficient actor-critic with experience replay.
\newblock \emph{ICLR}, 2016.

\bibitem[Wierstra et~al.(2007)Wierstra, Foerster, Peters, and
  Schmidhuber]{Wierstra2007}
Daan Wierstra, Alexander Foerster, Jan Peters, and J{\"u}rgen Schmidhuber.
\newblock Solving deep memory pomdps with recurrent policy gradients.
\newblock In \emph{Artificial Neural Networks -- ICANN 2007}, pages 697--706,
  Berlin, Heidelberg, 2007.

\bibitem[Zhang et~al.(2015)Zhang, Levine, McCarthy, Finn, and
  Abbeel]{Zhang2015Robo}
Marvin Zhang, Sergey Levine, Zoe McCarthy, Chelsea Finn, and Pieter Abbeel.
\newblock Policy learning with continuous memory states for partially observed
  robotic control.
\newblock \emph{CoRR}, abs/1507.01273, 2015.

\bibitem[Zhu et~al.(2017)Zhu, Li, and Poupart]{Zhu2017}
Pengfei Zhu, Xin Li, and Pascal Poupart.
\newblock On improving deep reinforcement learning for pomdps.
\newblock \emph{arxiv.org/abs/1704.07978}, 2017.

\end{thebibliography}
\footnotesize{
}

\newpage
\appendix
\setcounter{figure}{0}
\small
\section{Implementation}
We approximate transition density $p_\theta(z_t|h_{t-1}, a_{t-1})$, encoder $q_\phi(z_t|h_{t-1}, a_{t-1}, o_t)$ and decoder $p_\theta(o_t|z_t, h_{t-1}, a_{t-1})$ both as multivariate Gaussian distributions, each by two fully connected layers and two separate fully connected output layers for the mean and the variance respectively.

All $a$ and $o$ are encoded before passing into a network: $a$ is encoded by one fully connected layer of size $64$, and $o$ by two fully connected layers. Multiple inputs are concatenated as one vector. All covariance matrices are diagonal.

For actor-critic we use one fully connected layer followed by three output fully connected layers: one for the value function of size $1$, one for the policy mean and the other for the policy variance, both of size $2$. The behavior policy is also a fully connected layer followed by a layer for the mean and the other layer for variance.

To aggregate information in the particle filter, we first encode each weight by one fully connected layer of size $64$, and encode the corresponding particle value and the feature of weight into a belief feature by two fully connected layers. The $K$ belief features are then concatenated and encoded by a fully connected output layer in $b$ space. All network weights are shared across particles.

$z$, $h$ and $b$ all have a dimension of $128$. All other fully connected layers have a size of $128$ if not specified otherwise. RNNs are gated recurrent units (GRUs). All hidden layers are activated by rectified linear units (ReLUs). All network weights are initialized using orthogonal initializer. Batch normalization is not implemented.

We use RMSProp with shared statistics across threads as the optimizer which works well with RNNs. Momentum is not considered. The decay factor in RMSProp is $\alpha=0.99$. Gradients are truncated at the norm of $0.5$. The relative weights for $\mathcal{L}^{\theta,\phi,\kappa}_V$ and $\mathcal{L}^{\theta,\phi,\omega}_H$ 
are set to $0.5$ and $0.01$ as in A3C, the weight for $\mathcal{L}^{\theta, \phi}_\mathrm{ELBO}$ is $0.1$.

Our hyperparameters under tuning include learning rate $([1e-5, 1e-3]\textnormal{~log uniform})$ and regularization factor $\varepsilon$ in RMSProp $([1e-5, 1e-1]\textnormal{~log uniform})$.

\section{Patient Variables}
\begin{table}[htp!]
\begin{center}
\vspace{-0.3cm}
\caption{$48$ patient variables involved as observations. C: continuous/multi-valued, B: binary. \label{tb:patient_var}}
\scalebox{0.9}{
\begin{tabular}{ |m{3.8cm}|m{7cm}|m{0.9cm}| }
 \hline
 Category & Items & Type \\ 
 \hline
 \multirow{5}{*}{Demographics} & age (in days) & C \\
 & gender & B \\ 
 & weight (kg) & C \\ 
 & readmission to intensive care & B \\ 
 & elixhauser score (premorbid status) & C \\ 
 \hline
 \multirow{11}{*}{Vital signs} & SOFA (based on current step) & C \\
 & SIRS & C \\ 
 & Glasgow coma scale & C \\ 
 & heart rate & C \\ 
 & systolic blood pressure & C \\
 & diastolic blood pressure & C \\
 & mean blood pressure & C \\ 
 & shock index & C \\ 
 & respiratory rate & C \\ 
 & SpO$_2$ & C \\
 & temperature & C \\
 \hline
 \end{tabular}}
\end{center}
\end{table}
 
 \newpage
 \begin{table}[htp!]
\begin{center}
\vspace{-0.3cm}
\scalebox{0.9}{
\begin{tabular}{ |m{3.8cm}|m{7cm}|m{0.9cm}| }
 \hline
 \multirow{2}{*}{Ventilation parameters} & mechanical ventilation & B \\
 & FiO$_2$ & C \\ 
 \hline
 \multirow{26}{*}{Laboratory values} & potassium & C \\
 & sodium & C \\ 
 & chloride & C \\ 
 & magnesium & C \\ 
 & calcium & C \\ 
 & ionized calcium & C \\ 
 & carbon dioxide & C \\
 & glucose & C \\ 
 & BUN & C \\
 & creatinine & C \\
 & SGOT & C \\
 & SGPT & C \\ 
 & total bilirubin & C \\ 
 & albumin & C \\ 
 & hemoglobin & C \\ 
 & white blood cells count & C \\ 
 & platelets count & C \\
 & PTT & C \\ 
 & PT & C \\
 & INR & C \\
 & pH & C \\
 & PaO$_2$ & C \\ 
 & PaCO$_2$ & C \\ 
 & base excess & C \\ 
 & bicarbonate & C \\ 
 & lactate & C \\
 \hline
 \multirow{2}{*}{Fluid balance} & urine output (over $4$h) & C \\
 & accumulated fluid balance since admission & C \\ 
 \hline
 \multirow{2}{*}{Other interventions} & renal replacement therapy & B \\
 & sedation & B \\ 
 \hline
\end{tabular}}
\end{center}
\end{table}

\section{Cohort}
\subsection{Cohort Selection}
\begin{figure}[!htp]
	\centerline{
		\begin{tabular}{cc}
			\includegraphics[width=0.5\textwidth]{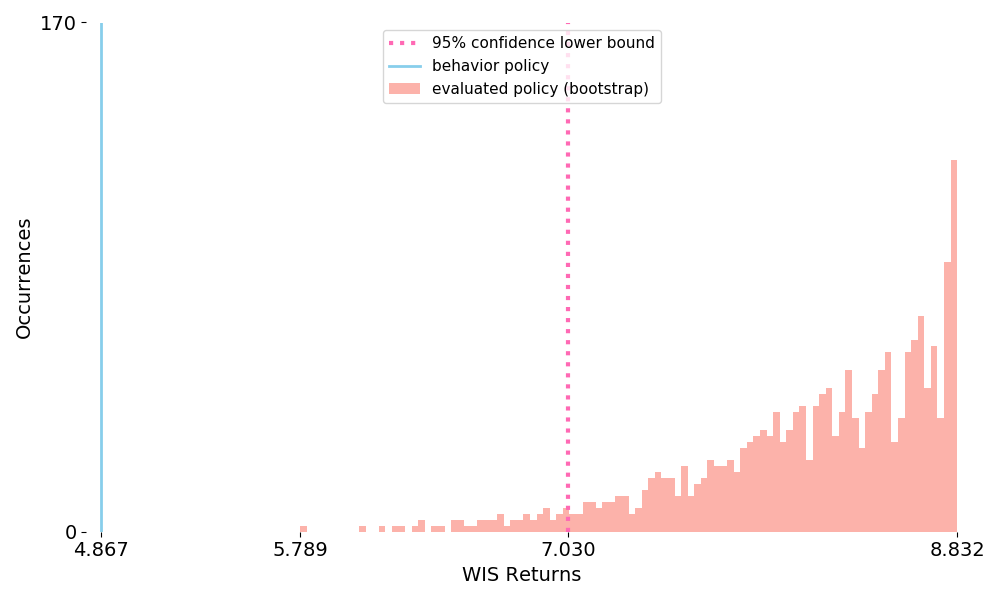}
			&
			\includegraphics[width=0.5\textwidth]{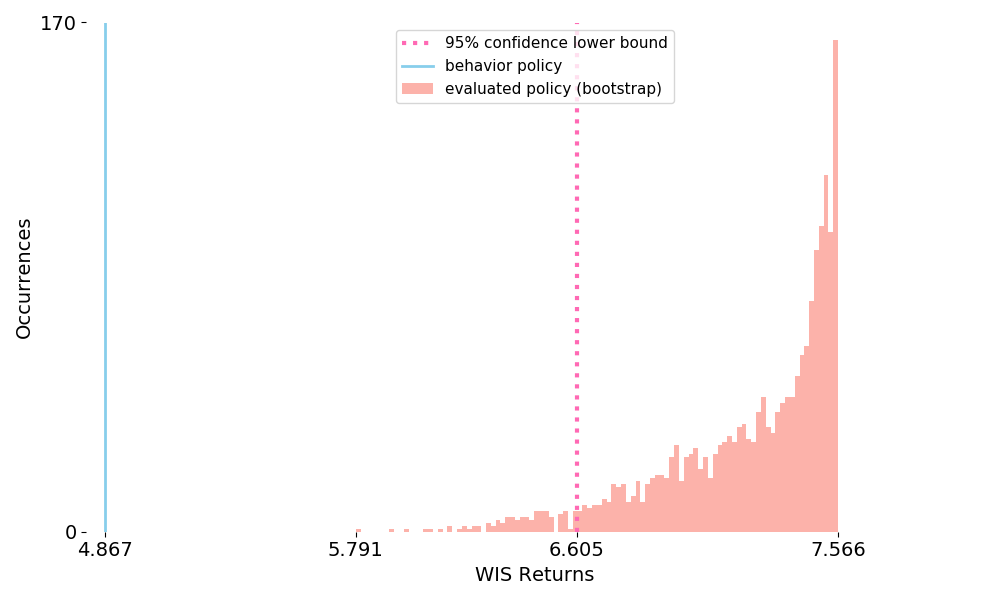}\\
			~(a)  & ~~(b)  \\
		\end{tabular}
	}
\vspace{-0.1cm}
\caption{Histograms on bootstrap samples of WIS return of test set. (a) when per-trajectory IS weights clipped to $[1e-30, 1e4]$. (b) when per-trajectory IS weights clipped to $[1e-20, 1e3]$}
\label{fig:histogram_oppe}
\end{figure}

Our cohort is a retrospective database, Medical Information Mart for Intensive Care Clinical Database (MIMIC-III) \cite{Johnson2017MAF} that contains de-identified health-related records of patients during their stays in a hospital ICU. We investigate the treatment optimization for sepsis, to which standard approach in current medical guidance lacks. We include data based on the same conditions as \cite{Komorowski2018naturemed}, selecting adult patients in the MIMIC-III database who conform to the international consensus sepsis-3 criteria \cite{Singer2016}, excluding admissions in which treatment was withdrawn or mortality was not documented. For all patients, data are extracted from up to $24$ hours preceding, and until up to $48$ hours ensuing the estimated onset of sepsis, to which the time series are aligned. Time series are temporally discretized with $4$-hour intervals, resulting in $18,914$ ICU admissions or $251,843$ steps in total. $2,000$ admissions are randomly sampled and excluded from training for evaluation.

\subsection{Data Extraction}
The treatments and patient variables to consider are also identical to those in \cite{Komorowski2018naturemed}, recounted here for completeness, with \textit{value continuities} retained. Our action space consists of the maximum dose of vasoppressors administered and the total volume of intravenous fluids injected over each $4$h time bin. The vasopressors include norepinephrine, epinephrine, vasopressin, dopamine and phenylephrine, and are converted to norepinephrine-equivalent with dose correspondence \cite{Brown2013vasopressor}. The intravenous fluids include boluses and background infusions of crystalloids, colloids and blood products, normalized by tonicity. Both actions are continuous. We model both $\pi$ and $\mu$ as Gaussian distributions. The output of each of the two policy networks contains two separate layers, one for the mean, one for the variance, both are two dimensional.

A table of the $48$ continuous or binary patient variables we assume to be of interest is given in Supp. Table \ref{tb:patient_var}. Notice that not all $48$ variables are available at each and all of the steps due to lower frequency or failure to record. We do \textit{not} interpolate those missing data, treating them instead as part of partial observability of the task.

The single goal in the POMDP is survival. Transitions to the terminal discharge without deceasing in $90$ days are rewarded by $+10$, those to the terminal death, either in-hospital or $90$ days from discharge, are penalized by $-10$. No intermediate reward is set so as to encourage learning from scratch.

\subsection{Data Preprocessing}
For measurements among the extracted $48$ patient variables with multiple records within a $4$h time bin, records are averaged (e.g. heart rate) or summed (e.g. urine output) as appropriate. All observation and action values are normalized between $[0, 1]$, action values for both drugs are then converted to $a\gets a^{0.3}$ to dilate small values because smaller dosages are selected more frequently by clinicians in respective range and we want to capture nuances.

\section{Further OPPE}
As the method for policy optimization is orthogonal to that for policy evaluation, and our major investigation is in the former, we simply clip each $W_n$ into $[W_-, W^-]$ to reduce variance. More truncation results in lower variance, at the cost of ending up evaluating an in-between policy that is less like $\pi$ but more like $\mu$.

We exemplify two sets of truncation thresholds, $[1e-30, 1e4]$ and $[1e-20, 1e3]$ to show the trend. To capture variance, we estimate WIS return of the test set through bootstrapping, resampling from the set $2,000$ times, resulting in $2,000$ bootstrap estimates. Each estimate is the result of applying WIS to the resampled trajectories. The histograms of the estimates are shown in Supp. Figure \ref{fig:histogram_oppe}, together with the $95\%$ confidence lower bound and the value of the behavior policy $R^\mu = \frac{1}{N}\sum_n^N R_n$. In both cases, the $95\%$ confidence lower bounds of $\pi$ value exceed the value of $\mu$ by considerable margins.

\newpage
\section{Pseudocode}
\begin{algorithm}
\DontPrintSemicolon 
$\mathcal{P}^*\gets\mathcal{P}_t, ~ b^*\gets b_t, ~d\gets0
\mathcal{F}\gets\o$

\For{$s=1,\dots,N_s$} {
    sample a subspace $\mathcal{A}$
    
sample ancestor indices $x^{1:n_z}$ from $\mathcal{P}^*$\;

\For{$a\in\mathcal{A}$}{
    \For{$i=1,\dots , n_z$}{
      sample $z^{d+1}_{a,i}\sim p_\theta(\cdot|h^{d,x^i}, a)$
      
      \For{$j=1,\dots , n_{z,o}$}{
      sample $o^{d+1}_{a,i,j}\sim p_\theta(\cdot|z^{d+1}_{a,i}, h^{d,x^i}, a)$
      
      $\mathcal{P}^{d+1}_{a,i,j}, b^{d+1}_{a,i,j} \gets \mathrm{UPDATE}(\mathcal{P}^*, a, o^{d+1}_{a,i,j})$
        }
      }
      $\Psi(b^*, a) \gets R(b^*, a)+ \frac{\gamma}{\eta^d(a)} \sum_{i=1}^{n_z}\sum_{j=1}^{n_{z,o}} \big( \frac{1}{K}\sum_{k=1}^K w^{d+1,k}_{a,i,j} \big) V_\kappa\big( b^{d+1}_{a,i,j} \big)$
    }
    $\hat{\pi}(a|b^*) \gets\frac{\exp\big( \Psi(b^*, a)/\beta \big)}{\int_\mathcal{A} \exp\big( \Psi(b^*, a')/\beta \big)\mathrm{d}a'}$
    
    $a_\mathcal{T}(b^*)\sim\hat{\pi}(\cdot|b^*)$
    
    $\mathcal{F}\gets\mathcal{F}\cup\big\{b^{d+1}_{i,j, a_\mathcal{T}(b^*)}, \forall i, j \big\}$
    
  \If{$s<N_s$}{
    $b^*\gets\mathop{\mathrm{arg\,max}}_{b \in\mathcal{F}} ~\gamma^{D(b)} \prod_{\substack{d=1\\
b^0\rightarrow b}}^{D(b)} \frac{1}{\eta^{d-1}(*)}\big( \frac{1}{K}\sum_{k=1}^K w^{d,k} \big)$

$d\gets$ depth of $b^*$

$\mathcal{P}^*\gets$ where $b^*$ is extracted
  }
}
$v_{\mathcal{T}}(b)\gets V_{\kappa}(b)~\forall~b\in\mathcal{F}$

\For{$d=\big(\mathop{\mathrm{max}}_{b \in\mathcal{F}}D(b)-1\big),\dots,0$}{
\For{$\mathrm{each ~expanded~} b^d$}{
$v_\mathcal{T}(b^d)\gets
R\big(b^d, *\big) +\frac{\gamma}{\eta^d(*)} \sum_{i=1}^{n_z}\sum_{j=1}^{n_{z,o}} \big( \frac{1}{K}\sum_{k=1}^K w^{d+1,k}_{*,i,j} \big) V_\kappa\big( b^{d+1}_{*,i,j} \big)$
}
}
\KwOut{$v_\mathcal{T}(b^0)$} \;
\caption{Heuristic search tree $\mathcal{T}\big( \mathcal{P}_t, ~b_t \big)$}
\end{algorithm}

\begin{algorithm}
\DontPrintSemicolon
initialize $\theta, ~\phi, ~\kappa, ~\omega, ~\zeta$

concatenate epochs of shuffled admission trajectories

$o_0\gets$ first observation

$\mathcal{P}_0, ~b_0\gets \mathrm{INIT}(o_0)$

$t\gets0$

\Repeat{$\mathrm{converged}$}{
$\mathcal{L}^{\theta, \phi}_\mathrm{ELBO}, ~\mathcal{L}^{\theta, \phi, \kappa}_V, ~\mathcal{L}^{\theta, \phi, \omega}_\pi, ~\mathcal{L}^{\theta, \phi, \omega}_H, \mathcal{L}^{\theta,\phi,\zeta}_{\mu} \gets0$

$t_\mathrm{start}\gets t$

\For{$\tau = t, \dots, (t+L-1)$}{
$\mathcal{L}^{\theta, \phi}_\mathrm{ELBO} -= \log \big( \frac{1}{K}\sum_k w^k_\tau \big)$

\For{$n=1,\dots,N_e \mathrm{~(compute~ in~ parallel)}$ }{
$v^n_\mathcal{T}(b_\tau)\gets\mathcal{T}\big( \mathcal{P}_\tau, ~b_\tau \big)$}
 $\mathcal{L}^{\theta, \phi, \kappa}_V += \big( \sum_n^{N_e}v_\mathcal{T}^n(b_\tau)/N_e -V_\kappa(b_\tau) \big)^2$
 
 \uIf{$\tau== \mathrm{last ~nonterminal ~step ~in ~current ~trajectory}$}{
 $\hat{A}^\infty\gets0$
 
 \For{$i = \tau,\dots, t_\mathrm{start}$}{
 $\delta_i \gets \rho_i \big( r_i+\gamma V_\kappa(b_{i+1})-V_\kappa(b_i) \big), ~\big( \delta_\tau\gets\rho_\tau(r_\tau - V_\kappa(b_\tau)) \big)$

$\hat{A}^\infty\gets \gamma c_i\hat{A}^\infty +\delta_i$

$\mathcal{L}^{\theta, \phi, \omega}_\pi -=\log\pi(a_i|b_i) \hat{A}^\infty$

$\mathcal{L}^{\theta, \phi, \omega}_H -=\int\pi(a|b_i)\log\pi(a|b_i)\mathrm{d}a$

$\mathcal{L}^{\theta,\phi,\zeta}_{\mu} -= \log \mu(a_i|b_i)$}
$t_\mathrm{start}\gets \tau+1$

$\mathcal{P}_{\tau+1}, ~b_{\tau+1}\gets \mathrm{INIT}(o_{\tau+1})$}

\Else{$\mathcal{P}_{\tau+1}, ~b_{\tau+1} \gets \mathrm{UPDATE} (\mathcal{P}_{\tau}, a_{\tau}, o_{\tau+1})$}
}

\For{$s=(t+L), \dots, \mathrm{~last ~nonterminal ~step ~in ~current ~trajectory}$}{$\_, ~b_s \gets \mathrm{UPDATE} (\mathcal{P}_{s-1}, a_{s-1}, o_s)$}

$\hat{A}^\infty\gets \sum_{s=t+L}^\infty\gamma^{s-t-L} \Big( \prod_{i=t+L}^{s-1}c_i \Big) \delta_s$

\For{$i=(t+L-1), \dots, t_\mathrm{start}$}{
$\delta_i \gets \rho_i \big( r_i+\gamma V_\kappa(b_{i+1})-V_\kappa(b_i) \big)$

$\hat{A}^\infty\gets \gamma c_i\hat{A}^\infty +\delta_i$

$\mathcal{L}^{\theta, \phi, \omega}_\pi -=\log\pi(a_i|b_i) \hat{A}^\infty$

$\mathcal{L}^{\theta, \phi, \omega}_H -=\int\pi(a|b_i)\log\pi(a|b_i)\mathrm{d}a$

$\mathcal{L}^{\theta,\phi,\zeta}_{\mu} -= \log \mu(a_i|b_i)$}
$\mathcal{L}\gets\sum\mathrm{rescale}\big( \mathcal{L}^{\theta, \phi}_\mathrm{ELBO}, ~\mathcal{L}^{\theta, \phi, \kappa}_V, ~\mathcal{L}^{\theta, \phi, \omega}_\pi, ~\mathcal{L}^{\theta, \phi, \omega}_H, \mathcal{L}^{\theta,\phi,\zeta}_{\mu} \big)/L$

update $\theta, \phi, \kappa, \omega, \zeta$ with $\nabla(\mathcal{L})$

$t\gets t + L$
}

\caption{AEHS}
\end{algorithm}

\newpage
\begin{algorithm}
\DontPrintSemicolon
\For{$k=1,\dots, K$}{
sample $h\sim p_{\theta}(h_0), ~a\sim \mathrm{random}$

sample $z^k_t\sim q_\phi(\cdot|h, a, o_0)$

$h^k_t\gets\Upsilon_\mathrm{RNN}(h, z_t^k, a, o_t|\theta)$

$w_t^k \gets \frac{p_\theta(z_t^k|h, a) p_\theta(o_0|z^k_t, h, a)}{q_\phi(z^k_t|h, a, o_0)}$

insert $\left\langle h_t^k, z_t^k, w_t^k \right\rangle$ into $\mathcal{P}_t$}
aggregate information in $\mathcal{P}_t$ into $b_t$

\KwOut{$\mathcal{P}_t, ~b_t$}
\caption{Belief state initialization $\mathrm{INIT}(o_0)$}
\end{algorithm}

\begin{algorithm}
\DontPrintSemicolon
\For{$k=1, \dots, K$}{
sample $x^k_{t-1}\sim w^{x^k_{t-1}}_{t-1}/\sum_i w^i_{t-1}$

sample $z^k_t\sim q_\phi(\cdot|h_{t-1}^{x^k_{t-1}}, a_{t-1}, o_t)$

$h^k_t\gets\Upsilon_\mathrm{RNN}(h_{t-1}^{x^k_{t-1}}, z_t^k, a_{t-1}, o_t|\theta)$

$w_t^k \gets \frac{p_\theta(z_t^k|h^{x^k_{t-1}}_{t-1}, a_{t-1}) p_\theta(o_t|z^k_t, h^{x^k_{t-1}}_{t-1}, a_{t-1})}{q_\phi(z^k_t|h_{t-1}^{x^k_{t-1}}, a_{t-1}, o_t)}$

insert $\left\langle h_t^k, z_t^k, w_t^k \right\rangle$ into $\mathcal{P}_t$}
aggregate information in $\mathcal{P}_t$ into $b_t$

\KwOut{$\mathcal{P}_t, ~b_t$}
\caption{Belief transition $\mathrm{UPDATE}\big( \mathcal{P}_{t-1}, a_{t-1}, o_t \big)$}
\end{algorithm}

\end{document}